\documentclass[letterpaper]{article} 
\usepackage{aaai24}  
\usepackage{times}  
\usepackage{helvet}  
\usepackage{courier}  
\usepackage[hyphens]{url}  
\usepackage{graphicx} 
\usepackage{natbib}  
\usepackage{caption} 
\usepackage{adjustbox}
\usepackage{multirow}
\usepackage{booktabs}
\usepackage{upgreek}
\usepackage{bm}
\usepackage{amsmath,amssymb,amsfonts}
\usepackage{xcolor}
\usepackage{makecell}

\usepackage{algorithm}
\usepackage{algpseudocode}

\usepackage{newfloat}
\usepackage{listings}
\DeclareCaptionStyle{ruled}{labelfont=normalfont,labelsep=colon,strut=off} 
\floatstyle{ruled}
\newfloat{listing}{tb}{lst}{}
\floatname{listing}{Listing}
\usepackage{pifont}
\newcommand{\cmark}{\ding{51}}
\newcommand{\xmark}{\ding{55}}

\title{PreRoutGNN for Timing Prediction with Order Preserving Partition: Global Circuit Pre-training, Local Delay Learning and Attentional Cell Modeling}

\author {
    Ruizhe Zhong\textsuperscript{\rm 1},
    Junjie Ye\textsuperscript{\rm 2},
    Zhentao Tang\textsuperscript{\rm 2},
    Shixiong Kai\textsuperscript{\rm 2},\\
    Mingxuan Yuan\textsuperscript{\rm 2},
    Jianye Hao\textsuperscript{\rm 2, 3},
    Junchi Yan\textsuperscript{\rm 1}\thanks{Correspondence author.}
}
\affiliations {
    \textsuperscript{\rm 1}Dept. of Computer Science and Engineering \& MoE Key Lab of AI, Shanghai Jiao Tong University\\
    \textsuperscript{\rm 2}Huawei Noah’s Ark Lab\\
    \textsuperscript{\rm 3}College of Intelligence and Computing, Tianjin University\\
    zerzerzerz271828@sjtu.edu.cn, kourenmu@163.com, tangzhentao1@huawei.com, kaishixiong@huawei.com, \\
    Yuan.Mingxuan@huawei.com, haojianye@huawei.com, yanjunchi@sjtu.edu.cn
}

\begin{document}

\maketitle
\begin{abstract}
Pre-routing timing prediction has been recently studied for evaluating the quality of a candidate cell placement in chip design. It involves directly estimating the timing metrics for both pin-level (slack, slew) and edge-level (net delay, cell delay), without time-consuming routing. However, it often suffers from signal decay and error accumulation due to the long timing paths in large-scale industrial circuits. To address these challenges, we propose a two-stage approach. First, we propose global circuit training to pre-train a graph auto-encoder that learns the global graph embedding from circuit netlist. Second, we use a novel node updating scheme for message passing on GCN, following the topological sorting sequence of the learned graph embedding and circuit graph. This scheme residually models the local time delay between two adjacent pins in the updating sequence, and extracts the lookup table information inside each cell via a new attention mechanism. To handle large-scale circuits efficiently, we introduce an order preserving partition scheme that reduces memory consumption while maintaining the topological dependencies. Experiments on 21 real world circuits achieve a new SOTA $R^2$ of 0.93 for slack prediction, which is significantly surpasses 0.59 by previous SOTA method. Code will be available at: \url{https://github.com/Thinklab-SJTU/EDA-AI}.
\end{abstract}

\section*{Introduction}
The process of integrated circuit (IC) design can be thought of as a series of hierarchical decomposition steps, typically consisting of architectural design, logic synthesis, physical design (macro \& cell placement, routing) and verification.
As IC becomes more and more complex, timing constraints, such as delay and slack, are becoming more difficult to satisfy during the physical design stage. 
`Shift-left'~\cite{zhou2022heterogeneous} suggests circuit constraints and performance should be considered in earlier stages of design flow, for instance, taking timing into consideration during standard cell placement stage~\cite{liao2022dreamplace}.
Timing metrics are critical to judge the performance of a design, but accurate timing information is only available after routing, which is one of the most time-consuming steps. 
Since repetitive routing in cell placement stage is unacceptable, many analytical placers~\cite{lu2015eplace, lin2019dreamplace} use half-perimeter wirelength as a surrogate of design timing quality.

Machine learning has been widely applied in EDA design flow, such as logic synthesis~\cite{EasySO}, placement~\cite{mirhoseini2021graph, cheng2021joint, cheng2022policy, lai2022maskplace, lai2023chipformer} and routing~\cite{du2023hubrouter}.
However, accurate timing information can be accessed only after routing stage, which is time-consuming.
To estimate pre-routing timing metrics, machine learning has been introduced for timing prediction, such as delay~\cite{yang2022pre}, wirelength~\cite{xie2021net2, yang2022versatile}, to guide timing-driven cell placement.

Among various timing metrics, slack~\cite{hu2014tau} is one of the most important yet challenging metric for prediction. 
Slack is the difference between Required signal Arrival Time (RAT) and actual Arrival Time (AT), and it is calculated on each pin.
Accurate slack prediction can be accessed only after routing stage, which is a time consuming process.
To avoid the time consuming routing stage, directly estimating timing information is attractive, which can be defined formally as follows: given the intermediate results after standard cell placement stage and before routing stage, we want to predict slack for all pins after routing stage.
Usually, RAT is provided in external files, so we mainly focus on the prediction of AT.
However, in large designs, the timing path can be long with hundreds of circuit elements, where signal decay and error accumulation problems exist, making slack prediction challenging. 
Besides, directly applying deep learning into large circuits are always faced with high peak GPU memory cost problem, making training process hard or even intractable.

\begin{figure}[!t]
    \centering
    \includegraphics[width=0.217\textwidth]{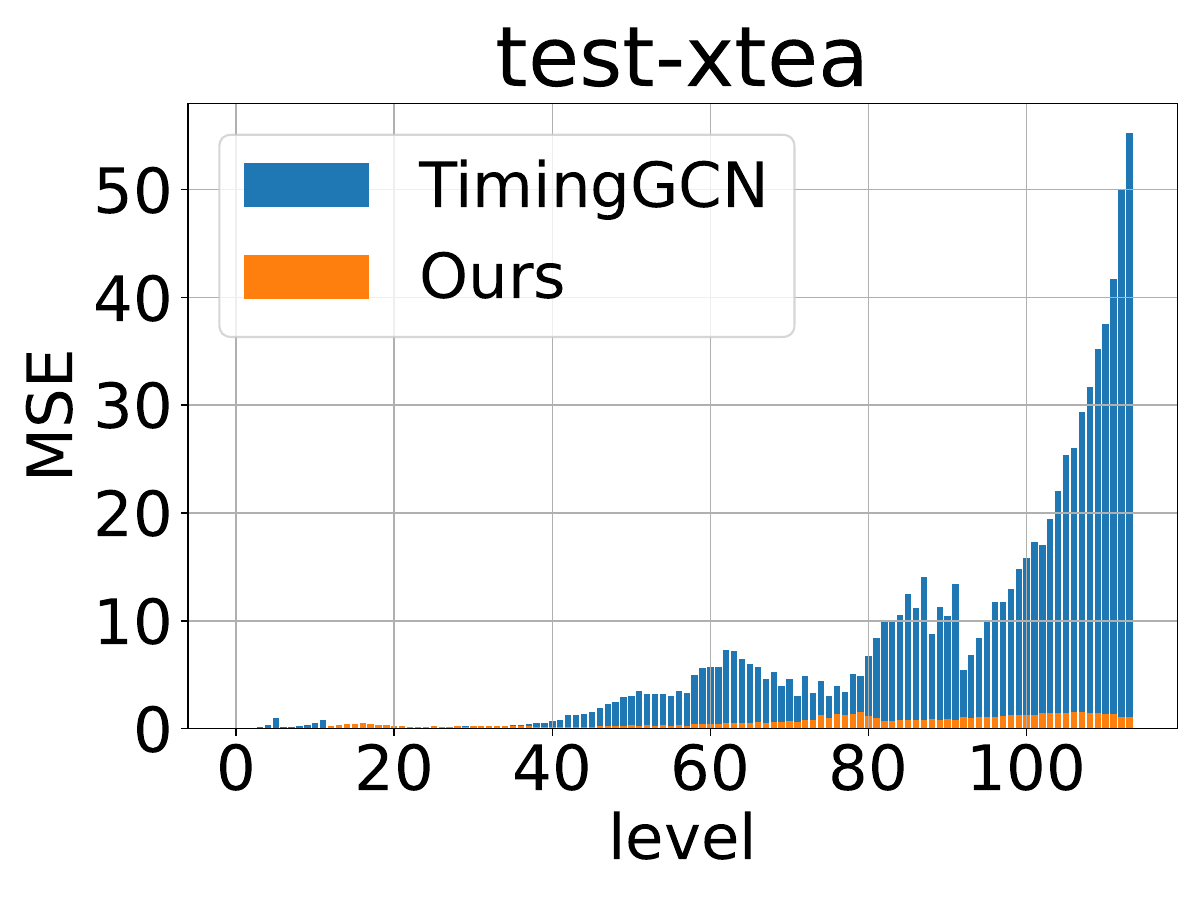}
    \includegraphics[width=0.243\textwidth]{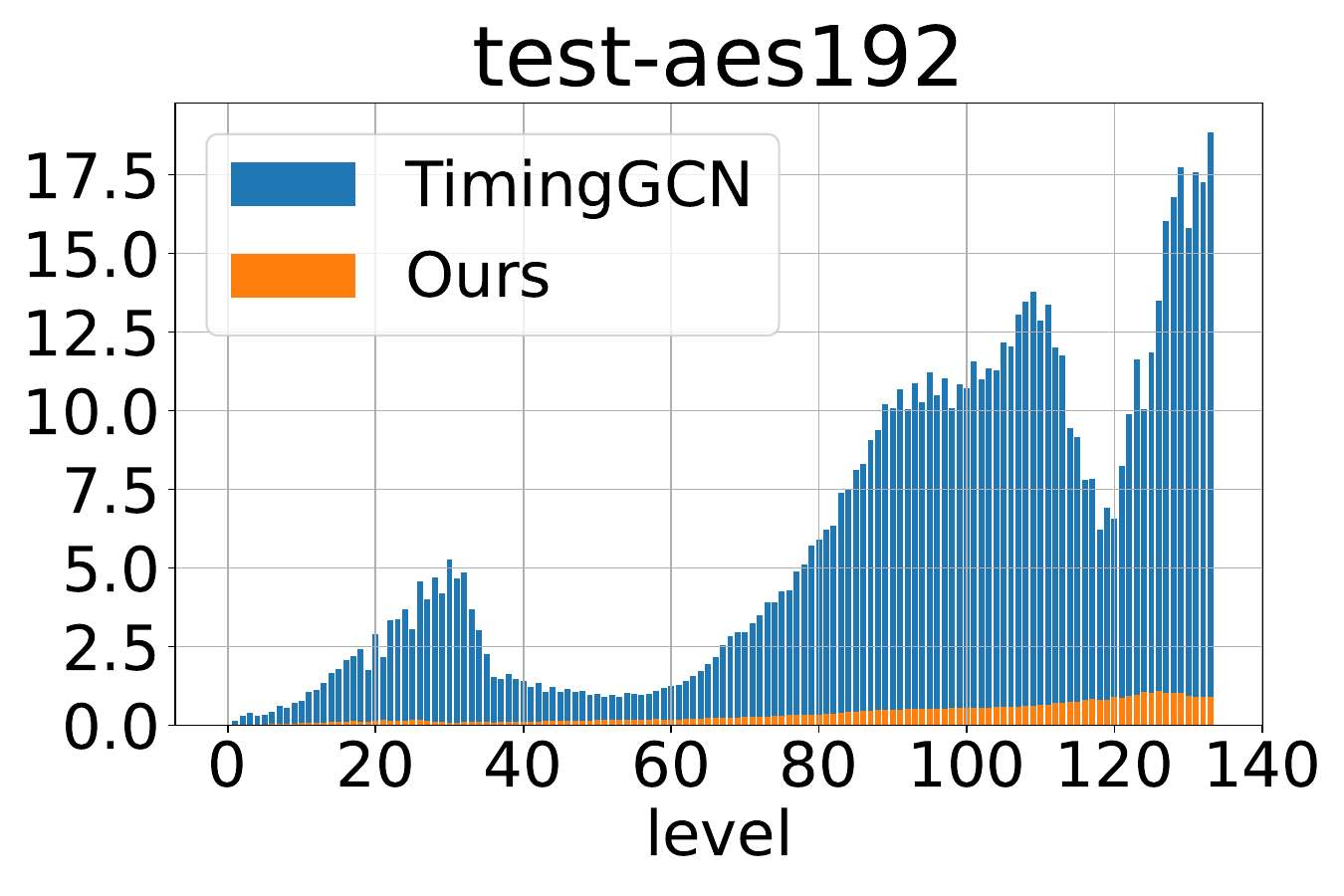}\\
    \caption{
    MSE of AT prediction v.s. topological level. 
    In previous ML-based methods TimingGCN~\cite{guo2022timing}, MSE increases rapidly with level, showing the existence of signal decay and error accumulation.
    }
    \label{fig:MSE-vs-level}
\end{figure}

Typical GNNs (Graph Neural Networks) such as GCN~\cite{kipf2016semi} and GAT~\cite{velivckovic2018graph} are designed to process graphs where local information is already sufficient. 
However, in circuits graphs, signals travel from primary inputs to timing endpoints, forming long timing paths~\cite{hu2014tau}, where long range dependencies and global view play a critical role but they are hard for typical GNN to handle. 
Previous methods for other timing metric (delay, wirelength) prediction~\cite{xie2018routenet, barboza2019machine, ghose2021generalizable, xie2021net2, yang2022pre, yang2022versatile} do not take the global graph information into consideration. 
For example, CircuitGNN~\cite{yang2022versatile} converts each circuit to a heterogeneous graph, and design the topo-geom message passing paradigm for prediction of wirelength and congestion. 
Wirelength is estimated in each net using half-perimeter wirelength, and congestion is predicted in local area measured with pixel values. 
Both tasks are enclosed in local area, where local information is already sufficient.
However, in pre-routing slack prediction, it relies all elements through timing paths, where long range dependencies and global view play a critical role, and they are absent in CircuitGNN.
Consequently, we cannot directly apply these GNNs into the prediction of slack.
Net$^2$~\cite{xie2021net2} for wirelength prediction applies graph partition and merges cluster embedding into node features, but cannot avoid the local receptive field enclosed in cluster. 
TimingGCN~\cite{guo2022timing} utilizes a timing engine inspired GNN to predict slack, where node embedding is updated with predecessors asynchronously.
However, it is prone to signal decay and error accumulation, demonstrated in Fig.~\ref{fig:MSE-vs-level}.
We first apply topological sorting~\cite{lasser1961topological} on circuit graph to get topological level (topological sorting order) of each pin. 
Then we calculate MSE (Mean Squared Error) between AT prediction and ground truth for pins belonging to the same sorting order. 
In other methods, MSE increases rapidly with level, demonstrating the existence of signal decay and error accumulation problems. 

Motivated by these issues, we propose our two-stage approach for pre-routing slack prediction.
In the first stage, we pre-train an auto-encoder to extract node and graph embeddings for following downstream tasks, providing a global receptive field and an informative representation.
In the second stage, we design a new message passing and node updating scheme for long timing paths, and train our GNN for pre-routing slack prediction.
We list features of different methods in Table~\ref{tab:model-feature-cmp}.
\begin{table}[!tb]
    \centering
    
    {\fontsize{9pt}{10pt}\selectfont
        \begin{tabular}{l|ccccccccccc}
            \toprule
             Model & Update scheme & Global & Embed. & Level  \\
             \midrule
             GCN 
             & Sync./Direct  & \xmark & \xmark & \xmark  \\
             GAT
             & Sync./Direct  & \xmark & \xmark & \xmark  \\
             GINE
             & Sync./Direct  & \xmark & \xmark & \xmark  \\
             GCNII
             & Sync./Direct  & \xmark & \xmark & \xmark  \\
             CircuitGNN
             & Sync./Direct  & \xmark & \xmark & \xmark  \\
             TimingGCN
             & Async./Direct & \cmark & \xmark & \xmark  \\
             Net$^2$
             & Sync./Direct & \cmark & \cmark & \xmark  \\
             LHNN
             & Sync./Direct & \xmark & \xmark & \xmark  \\
             Ours & Async./Residual & \cmark & \cmark & \cmark  \\
             \bottomrule
        \end{tabular}
    }
    \caption{
    Features comparison among different methods. 
    Residual means ours node updating scheme in Eq.~\ref{eq:residual-message-passing-paradigm}, which explicitly models net delay and cell delay.
    Embed. means graph embeddings.
    Level means we introduce topological sorting order into GNN. 
    }
    \label{tab:model-feature-cmp}
    
\end{table}
\textbf{The highlights of this work are:}
\begin{itemize}

\item \textbf{First time to introduce global circuit pre-training into timing (specifically slack) prediction. } 
Global view plays a critical role in addressing the signal decay and error accumulation issues.
We capture this feature via graph embeddings, which can be effectively learned by  global circuit  pre-training technique. 
Besides, our global circuit pre-training can serve as a plug-and-play module for other timing prediction GNN.

\item \textbf{Residual local learning of signal delay.} We design a new message passing and node updating scheme, with residual local learning to explicitly model local signal delay in long time paths.

\item \textbf{Attention-based cell modeling.}
We re-explore the modeling process of cell and explain it as a `query-index-interpolation' procedure. 
Based on this, we propose multi-head joint attention mechanism to depict it.

\item \textbf{Handling large-scale and strong performance.} 
For scalability and making training on large-scale circuits tractable, we design a order preserving graph partition algorithm to reduce memory cost while preserving the order dependencies.
Experiments on public benchmark with 21 real world circuits demonstrate the effectiveness of our method. 
For $R^2$ of slack prediction, we achieve new SOTA 0.93, surpassing 0.59 of peer SOTA method. 
\end{itemize}

\begin{figure*}[!t]
    \centering
    \includegraphics[width=1.0\linewidth]{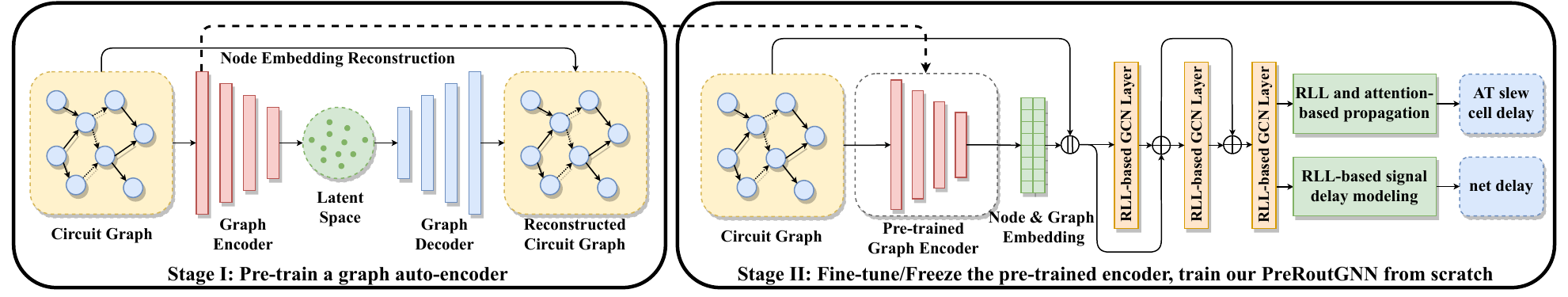}
    \caption{Pipeline of our approach. Circuit graph is represented as a heterogeneous DAG. We first implement training an auto-encoder by global circuit reconstruction in a self-supervised way. 
    After global circuit pre-training, we drop the decoder, and freeze/fine-tune the encoder to map each circuit graph to a low-dimensional latent space. 
    The latent vector is treated as the global graph embedding and concatenates to original node features. 
    Finally, heterogeneous circuit DAG is feed forward to the our PreRoutGNN for timing prediction. 
    }
    \label{fig:pipeline}
\end{figure*}

\section*{Preliminaries and Related Works}
\subsubsection{Static Timing Analysis}
Static timing analysis (STA) provides a profile of a design's performance by evaluating the timing propagation from inputs to outputs, forming multiple timing paths.
A timing path~\cite{hu2014tau} is a set of directed connections through circuit elements and connections, and all signals travel through it. 
After routing, STA should be done under four corner conditions: early-late/rise-fall (EL/RF)~\cite{hu2014tau}.

During STA, the circuit is represented as an Activity-On-Edge network, where each pin is a node and pin-to-pin connection is an edge. 
The activity cost on each edge is either net delay or cell delay. 
Starting from primary inputs (source nodes), the instant when a signal reaches a pin is quantified as the arrival time (AT). 
Similarly, starting from the primary outputs (sink nodes), the limits imposed for each AT to ensure proper circuit operation is defined as the required arrival time (RAT). 
The slack is defined as the difference between RAT and AT as follows:
\begin{equation}
    \label{eq:slack}
    \scalebox{1}{$slack^E  = AT^E - RAT^E, slack^L = RAT^L - AT^L,$}
\end{equation}
where $E,L$ means early and late, and a positive slack means timing constraint is satisfied. 
Since RAT is usually provided in external files, so we mainly focus the estimation of AT, and use provided RAT and estimated AT to calculate slack.

Besides, the signal transition is characterized by its slew, which is defined as the required amount time of transition between electrical levels.
Slew is also an important timing metric to evaluation the quality of circuit, and accurate slew can be accessed by STA after routing.

\subsubsection{Pre-training in EDA}
As training from scratch requires much time and data, pre-training has been recently and widely applied in vision~\cite{bao2022beit} and language~\cite{devlin2018bert, brown2020language}, and it is becoming a promising trend in EDA. 
Using pre-trained model not only saves these costs but provides more informative representation. 
DeepSeq~\cite{khan2023deepseq}, DeepGate~\cite{li2022deepgate} and Bulls-Eye~\cite{chowdhury2022too} transform circuits into AND-Inverter-Graph format and learn corresponding representation for logic synthesis. 
DeepPlace~\cite{cheng2021joint} uses reinforcement learning agent pre-trained on macro connection to jointly solve placement and routing.
However, few works about pre-training have been done in timing prediction, especially slack.
We think pre-training can provide a graph embedding containing global view, which plays a critical role in addressing the signal decay and error accumulation in long timing paths.

\subsubsection{GNNs for Timing Prediction}
GNNs are optimizable transformations for graph-structured data attributes. 
As they continue to evolve, new variants are emerging that target graph topological structure~\cite{zhao2023graphglow} and large-scale characteristics~\cite{wu2022nodeformer, wu2023difformer, wu2023simplifying}.
Due to the graph structure of circuit netlist, it is natural to apply GNN in EDA field. 
GCN~\cite{defferrard2016convolutional, kipf2016semi} is proposed as a generalization from low-dimensional regular grids to high-dimensional irregular domains. 
GAT~\cite{velivckovic2018graph} introduces attention mechanism to assign learnable weights for neighbours. 
GCNII~\cite{chen2020simple} introduces initial residual and identity mapping to mitigate over-smoothing problem. 
GINE~\cite{hu2019strategies} is a variant of GIN~\cite{xu2019powerful} with additional edge features. At first, it is designed to address the graph isomorphism problem, but it also demonstrates strong ability in other graph related tasks, e.g. node classification. 
However, aforementioned GNNs are not specifically designed for circuits containing long timing paths, thus they are faced with local receptive field and over-smoothing problems. 
CircuitGNN~\cite{yang2022versatile} converts each circuit to a heterogeneous graph, and designs the topo-geom message passing paradigm for prediction of wirelength and congestion.
TimingGCN~\cite{guo2022timing} is a timing engine inspired GNN especially for pre-routing slack prediction.
LHNN~\cite{wang2022lhnn} uses lattice hyper-graph neural network to prediction routing congestion.
Net$^2$~\cite{xie2021net2} applies graph attention network into pre-placement net length estimation.

\section*{Methodology}
\textbf{Formulation.} 
Given the intermediate design result after placement and before routing, we predict the AT and slack for all pins in the circuit after routing.
Each circuit is represented as a heterogeneous and directed acyclic graph (DAG), containing one type of node and three types of edge. 
Primary inputs (input ports), primary outputs (output ports), and pins are treated as one type of node.
Cell, net and net\_inv are three types of edges, corresponding to fan-in to fan-out, net driver to net sink, net sink to net driver connection, respectively. 
Pipeline of our approach is shown in Fig.~\ref{fig:pipeline}.

\subsection*{Global Circuit Pre-training}
Due to the local receptive field of common message passing paradigm based GNNs, the global view is often absent.
It plays an essential role in accurately modeling the timing feature along timing path.
As a consequence, the lack of global view leads to the signal decay and error accumulation problems.
Motivated by this issue, we capture this feature via graph embeddings, which can be effectively learned by global circuit  pre-training.

The first stage of our approach is to train an auto-encoder for graph reconstruction in a self-supervised way. 
We use it to explicitly extract the global embedding for downstream tasks. 
It provides a global view for the whole graph, which is absent in previous methods. 
Since the pre-training is implemented in a self-supervised way, circuits without timing labels (AT, slew, delay) can also be used for graph reconstruction, saving the cost during complete design flow, e.g. routing and STA,  to access accurate labels.

The total loss of training auto-encoder is as follows:
\begin{equation}
\label{eq:loss-for-VAE}
{
\small
\mathcal{L}_{AE} = \lVert \mathbf{X} - D_\varphi\left(E_\theta(g, \mathbf{X})\right) \rVert_2^2+ \lambda \mathcal{L}_{KL},
}
\end{equation}
where $g$ is the circuit graph, $\mathbf{X}$ is the node feature.
To avoid arbitrarily high variance in latent space, we add a regularization term, a slight Kullback-Leibler divergence penalty towards a standard normal distribution, where $\lambda$ is the KL-penalty weight.
We train auto-encoder and our GNN with the same data, but timing labels do not involve this pre-training stage.
After that, we drop the decoder and freeze/fine-tune the encoder to get the node embeddings in latent space. 
The global graph embedding is extracted through a global average pooling layer. 
We concatenate these embeddings to original node features and feed them to following GNN.

\begin{figure}[!t]
    \centering
    \includegraphics[width=0.8\linewidth]{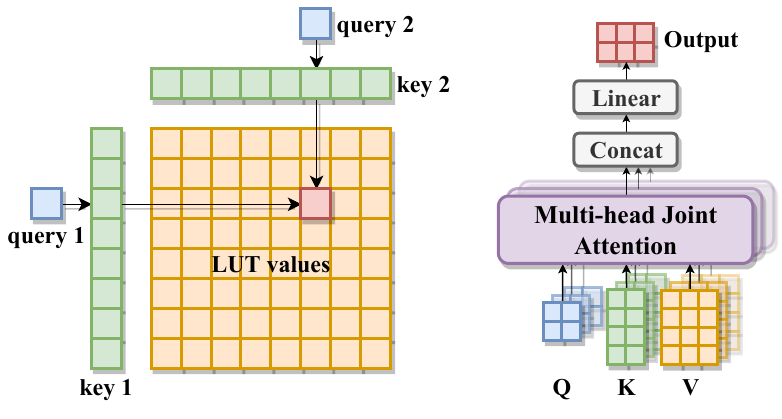}
    \caption{
        Multi-head Joint Attention (MJA) for cell modeling.
        To model this querying-indexing-interpolation process, we model it with a joint attention.
    }
    \label{fig:MJA}
\end{figure}

\subsection*{Residual Local Learning of Signal Delay}
In line with TimingGCN~\cite{guo2022timing},  we define \emph{topological level (level for short)} as a sub-set of whole graph's vertex set, and all vertices in a level have the same topological sorting order.
Each vertex belongs to one and only one level.
Given a level, we can set the sorting order of any its vertices as its level index, and this level index is unique.

From Fig.~\ref{fig:MSE-vs-level} we can see that, in previous estimation methods, MSE increases rapidly with topological level, suggesting the existence of signal decay and error accumulation. 
To mitigate these problems, we propose our message passing and node updating scheme for long timing paths.

Considering a circuit and its signal propagation, e.g. the computation of AT, the circuit can be represented as an Activity-On-Edge network~\cite{hu2014tau}, where the activity cost on each edge is net delay or cell delay. 
The AT of current node $i$ is computed as $AT_i = \max_{j \in \mathcal{N}(i)} \left ( AT_{j} + d_{ji} \right)$, where $\mathcal{N}(i)$ is the predecessors of $i$ and $d_{ji}$ is the delay from $j$ to $i$.

To explicitly model this delay, we introduce residual local learning (RLL) in our node updating scheme:
\begin{equation}
\label{eq:residual-message-passing-paradigm}
{
\begin{aligned}
 \mathbf{m}_e^{(t+1)} &= \mathbf{x}_u^{(t)} +  \phi \left (  \mathbf{x}_u^{(t)}, \mathbf{x}_v^{(t)}, \mathbf{f}_e^{(t)} \right), e \in \mathcal{E}  \\
 \mathbf{x}_{v}^{(t+1)} &=  \mathbf{x}_v^{(t)} + \rho \left( \left\{  \mathbf{m}_e^{(t+1)}: e  \in \mathcal{E}\right\} \right), 
\end{aligned}
}
\end{equation}
where $\mathbf{x}_v \in \mathbb{R}^{d_1}$ is the feature for node $v$, $e = (u,v)$ is the directed edge from $u$ to $v$, $\mathbf{f}_{e} \in \mathbb{R}^{d_2}$ is the feature for edge $e$, $\mathcal{E}$ is the edge set, $t$ is the GNN layer index, $\phi$ is the message function, $\rho$ is the reduce function, and $\psi$ is the update function.
This inductive bias better depicts the signal computation process and learning incremental value is the direct manifestation of learning cell delay and net delay.

\begin{figure}[!t]
    \centering
\includegraphics[width=1.0\linewidth]{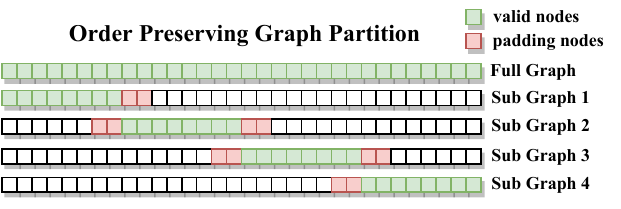}
    \caption{
    Illustration for order preserving graph partition. 
    Padding nodes are not involved in loss computing and BP. 
    }
    \label{fig:graph-partition}
\end{figure}

\textbf{Topological level encoding.}
Node topological sorting order is useful to address the signal decay and error accumulation, as it indicates the node location in timing path.
We incorporate topological level encoding to map the level index to a multi-frequency vector. 
Inspired by NeRF~\cite{mildenhall2021nerf} and Transformer~\cite{vaswani2017attention}, we use following encoding:
\begin{equation}
\label{eq:level-encoding}
\begin{aligned}
\scalebox{0.85}{$\gamma(x) = [ x, \sin(2^0 \uppi x/L), \cos(2^0  \uppi x/L),\sin(2^1 \uppi  x/L),$} \\
\scalebox{0.85}{$\cos(2^1  \uppi x/L), \ldots, \sin(2^{N-1} \uppi  x/L), \cos(2^{N-1} \uppi  x/L) ],$}
\end{aligned}
\end{equation}
where $x$ is the level index of each pin, $N$ is the number of different frequency components and $L$ is the maximum level. 
This level encoding explicitly incorporate high-frequency components into node embedding, assisting the GNN to identify the level information. 
Besides, the multi-head joint attention we proposed in the following section can be explained as a low-pass filter~\cite{wang2022anti, park2022vision}
The level encoding enhances and amplifies the high-frequency components, keeping them active during forward propagation.
As a result, it is beneficial to model the topological order.

\subsection*{Multi-head Joint Attention for Cell Modeling}
The output slew and cell delay for cell can be represented by the non-linear delay model~\cite{hu2015tau} stored in two-dimensional look-up tables (LUTs) in external files, e.g. Liberty files. 
The row and column indices of LUTs are pre-defined input slew and driving load. 
The values of LUTs are pre-defined cell delay or output slew.
In STA, given an input slew and a driving load, the corresponding output slew or cell delay can be calculated by referencing the table with two input indices and applying bi-linear interpolation, which is a `query-index-interpolation' procedure.

From the perspective of attention mechanism, input slew and driving load can be treated as queries, LUT's row and column as keys, and table of LUT as values. 
Since the two kinds of keys are not independent, we use multi-head joint attention (MJA) to model cell as follows:
\begin{equation}
\label{eq:MJA}
{
\small
    \begin{aligned}
    & \mathbf{K}  = \mathbf{K}_1 \times \mathbf{K}_2 \in \mathbb{R}^{n \times n \times d_k} \stackrel{\mathrm{reshape}}{\longrightarrow} \mathbb{R}^{n^2 \times d_k}\\
    & \mathbf{O} = \mathrm{softmax}\left( \frac{(\mathbf{QW}_Q)(\mathbf{KW}_K)^{\top}}{\sqrt{d}} \right) \mathbf{VW}_V,
    \end{aligned}
}
\end{equation}
where $m, n^2$ is the number of queries and keys/values respectively, $\mathbf{Q}  \in \mathbb{R}^{m \times d_q}, \mathbf{K}_1, \mathbf{K}_2  \in \mathbb{R}^{n \times d_k}, \mathbf{V}  \in \mathbb{R}^{n^2 \times d_v}$,  $\mathbf{K}_{ijk} = (\mathbf{K}_1)_{ik} (\mathbf{K}_2)_{jk}$ is the joint key matrix. 
$\mathbf{K}_1, \mathbf{K}_2$ are key matrices from LUT rows and columns, $\mathbf{Q}$ is value matrix from LUT value, and $\mathbf{Q}$ is the query matrix.
Our MJA is demonstrated in Fig.~\ref{fig:MJA}.
We calculate AT for node $i$ as follows if it is the end point of cell edge (edge type is `cell'):
\begin{equation}
    {
        \small
        \begin{aligned}
        \mathbf{Q}_{ji} & = \mathrm{MLP}(\mathbf{f}_i || \mathbf{f}_j || \mathbf{AT}_j) + \mathbf{AT}_j \\
        \mathbf{O}_{ji} & = \mathrm{MJA} ( \mathbf{Q}_{ji}, \mathbf{K}_{ji}, \mathbf{V}_{ji} ) \\
        \mathbf{{Y}}_{ji} & = \mathrm{LayerNorm}(\mathbf{O}_{ji} + \mathbf{Q}_{ji}) \\
        \mathbf{m}_{ji} & = \mathrm{MLP}(\mathbf{f}_i || \mathbf{f}_j || \mathbf{AT}_j || \mathbf{{Y}}_{ji}) + \mathbf{AT}_j \\
        \mathbf{a}_i & = \frac{\sum_{j \in \mathcal{N}(i)} \mathbf{m}_{ji}}{\mathrm{size}(\mathcal{N}(i))}, \mathbf{b}_i = \max_{j \in \mathcal{N}(i)} \mathbf{m}_{ji} \\
        \mathbf{AT}_i & = \mathrm{MLP}(\mathbf{a}_i || \mathbf{b}_i || \mathbf{f}_i).
        \end{aligned}
    }
\end{equation}

\begin{algorithm}[!t]
    \caption{Order Preserving Graph Partition}
    \label{alg:order-preserving-graph-partition}
    \begin{algorithmic}[1] 
        \Require Circuit directed acyclic graph $G$, maximum size $m$ for sub-graph, number of padding levels $k$.
        \Ensure All partitioned sub-graphs $G_s$ of $G$.

        \Function{TopologicalSort}{$G$}
           \State $L$ = [\,\,];
           \State {\color{gray}\# $l$ is a list of nodes belonging to the same level}
            \For{level $l$ in $G$} 
                \State $L$.append($l$);
            \EndFor
            \State \textbf{return} $L$
        \EndFunction

        \State $L$ = \textproc{TopologicalSort}($G$);
        \State $i=0, j=0, g_s=$[\,\,]$, G_s = $[\,\,];
        \While{$i < \mathrm{len}(L)$}
        \If{$i < \mathrm{len}(L)$ \textbf{and} $\mathrm{len}(g_s) + \mathrm{len}(L[i]) < m$}
        \State Add all nodes in $L[i]$ to $g_s$; \quad $i = i + 1$;
        \Else
        \State {\color{gray}\# Padding nodes do not involve training}
        \State $g_s = L[\max(0,j-k):j] \,||\,    g_s\,    ||\,   L[i:\min(i+k, \mathrm{len}(L))]$;
        \State $j=i$;
        \State Construct sub-graph with nodes $g_s$ from $G$;
        \State Add the sub-graph to $G_s$; 
        \State $g_s = $ [\,\,];
        \EndIf
        
        \EndWhile
    \end{algorithmic}
\end{algorithm}

\subsection*{Order Preserving Graph Partition}
Feeding the entire graph into GNN for training is faced with huge peak GPU memory cost.
An appropriate graph partition method is necessary to reduce the memory cost while reserving topological order information. 
 
To adapt to the signal propagation, we propose a order preserving graph partition algorithm. 
Given a circuit graph, we first apply topological sorting~\cite{lasser1961topological} to compute the sorting order of each pin. 
After sorting, we construct sub-graphs by selecting continuous levels of pins based on the given maximum sub-graph size. 
However, the marginal nodes of sub-graph lose their $1 \sim k$ hop ($1$ to $k$) neighbours, and the inner nodes also lose $2 \sim k, 3 \sim k, \ldots, k$-th hop neighbours, where $k$ is the number of stacked GNN layers.  
To resolve this problem, we pad the sub-graph with $k$ preceding and successive levels of pins. 
Note that these padding nodes are excluded in loss computation and backward propagation (BP), which ensures nodes involved gradient descent have full access to their $k$-hop neighbours like in the whole graph. 
Detailed graph partition algorithm is shown in Alg.~\ref{alg:order-preserving-graph-partition} and Fig.~\ref{fig:graph-partition}. 
Our partition algorithm reduces the GPU memory cost from more than 48 GB (out of memory) to 22 GB, and preserves the consistency without losing neighbours.

\begin{table*}[!t]
    \centering
    
    {\fontsize{9pt}{10pt}\selectfont
    \begin{tabular}{ll|ccccccccccccccc}
        \toprule

\multicolumn{2}{c|}{\multirow{2}{*}{Dataset}} & \multicolumn{5}{c}{$R^2_{uf}$ ($\uparrow$)}\\
\multicolumn{2}{c|}{} & GCNII & GAT & GINE & TimingGCN & PreRoutGNN\\
\midrule
\multirow{14}{*}{train} & BM64 & 0.763 $\pm$ 0.101 & 0.354 $\pm$ 0.223 & 0.623 $\pm$ 0.161 & 0.668 $\pm$ 0.252 & \textbf{0.986 $\pm$ 0.006}\\
~ & aes128 & 0.458 $\pm$ 0.207 & -0.245 $\pm$ 0.340 & 0.029 $\pm$ 0.352 & -0.055 $\pm$ 0.312 & \textbf{0.822 $\pm$ 0.008}\\
~ & aes256 & 0.581 $\pm$ 0.151 & -0.097 $\pm$ 0.306 & 0.067 $\pm$ 0.311 & 0.499 $\pm$ 0.102 & \textbf{0.937 $\pm$ 0.023}\\
~ & aes\_cipher & 0.020 $\pm$ 0.447 & -0.438 $\pm$ 0.427 & -0.162 $\pm$ 0.242 & 0.644 $\pm$ 0.123 & \textbf{0.967 $\pm$ 0.000}\\
~ & blabla & 0.889 $\pm$ 0.073 & 0.679 $\pm$ 0.131 & 0.737 $\pm$ 0.122 & 0.958 $\pm$ 0.013 & \textbf{0.997 $\pm$ 0.001}\\
~ & cic\_decimator & 0.719 $\pm$ 0.112 & 0.597 $\pm$ 0.163 & 0.589 $\pm$ 0.107 & 0.963 $\pm$ 0.014 & \textbf{0.996 $\pm$ 0.000}\\
~ & des & 0.924 $\pm$ 0.056 & 0.566 $\pm$ 0.225 & 0.776 $\pm$ 0.220 & 0.905 $\pm$ 0.058 & \textbf{0.993 $\pm$ 0.001}\\
~ & genericfir & 0.710 $\pm$ 0.122 & -0.436 $\pm$ 0.733 & 0.503 $\pm$ 0.219 & 0.222 $\pm$ 0.123 & \textbf{0.871 $\pm$ 0.013}\\
~ & picorv32a & 0.672 $\pm$ 0.096 & 0.239 $\pm$ 0.240 & 0.425 $\pm$ 0.169 & 0.839 $\pm$ 0.045 & \textbf{0.974 $\pm$ 0.007}\\
~ & salsa20 & 0.814 $\pm$ 0.066 & 0.530 $\pm$ 0.170 & 0.717 $\pm$ 0.073 & 0.792 $\pm$ 0.118 & \textbf{0.994 $\pm$ 0.002}\\
~ & usb & 0.747 $\pm$ 0.088 & 0.678 $\pm$ 0.115 & 0.607 $\pm$ 0.118 & 0.934 $\pm$ 0.035 & \textbf{0.996 $\pm$ 0.000}\\
~ & usb\_cdc\_core & 0.793 $\pm$ 0.095 & 0.502 $\pm$ 0.212 & 0.563 $\pm$ 0.177 & 0.961 $\pm$ 0.008 & \textbf{0.997 $\pm$ 0.001}\\
~ & wbqspiflash & 0.742 $\pm$ 0.091 & 0.613 $\pm$ 0.131 & 0.522 $\pm$ 0.136 & 0.956 $\pm$ 0.013 & \textbf{0.994 $\pm$ 0.002}\\
~ & zipdiv & 0.822 $\pm$ 0.070 & 0.703 $\pm$ 0.129 & 0.593 $\pm$ 0.188 & 0.971 $\pm$ 0.011 & \textbf{0.998 $\pm$ 0.001}\\
\midrule
\multirow{7}{*}{test} & aes192 & -0.017 $\pm$ 0.188 & -0.727 $\pm$ 0.173 & -0.048 $\pm$ 0.352 & 0.384 $\pm$ 0.205 & \textbf{0.937 $\pm$ 0.009}\\
~ & jpeg\_encoder & 0.056 $\pm$ 0.080 & -2.859 $\pm$ 0.667 & -0.285 $\pm$ 0.198 & 0.478 $\pm$ 0.131 & \textbf{0.764 $\pm$ 0.010}\\
~ & spm & -0.247 $\pm$ 0.309 & -0.474 $\pm$ 0.134 & 0.040 $\pm$ 0.161 & 0.802 $\pm$ 0.083 & \textbf{0.985 $\pm$ 0.006}\\
~ & synth\_ram & -0.913 $\pm$ 0.031 & -1.078 $\pm$ 0.025 & -1.007 $\pm$ 0.162 & 0.735 $\pm$ 0.079 & \textbf{0.977 $\pm$ 0.008}\\
~ & usbf\_device & -0.215 $\pm$ 0.065 & -0.946 $\pm$ 0.146 & -0.133 $\pm$ 0.094 & 0.347 $\pm$ 0.119 & \textbf{0.949 $\pm$ 0.002}\\
~ & xtea & 0.210 $\pm$ 0.044 & -0.055 $\pm$ 0.091 & 0.099 $\pm$ 0.181 & 0.888 $\pm$ 0.062 & \textbf{0.985 $\pm$ 0.000}\\
~ & y\_huff & -0.235 $\pm$ 0.450 & -0.481 $\pm$ 0.114 & -1.034 $\pm$ 0.999 & 0.490 $\pm$ 0.146 & \textbf{0.931 $\pm$ 0.029}\\
\midrule
\multicolumn{2}{c|}{Avg. train} & 0.689  & 0.303  & 0.471  & 0.733  & \textbf{0.966}\\
\multicolumn{2}{c|}{Avg. test} & -0.195  & -0.946  & -0.338  & 0.589  & \textbf{0.933}\\

\bottomrule
\end{tabular}
}
\caption{
Slack prediction with $R_{uf}^2$ as primary metric. 
Experiments are implemented on 5 different seeds.
    }
\label{tab:main-results-random-seeds}
\end{table*}

Finally, we build our \textbf{PreRoutGNN} (\textbf{Pre}-\textbf{Rout}ing estimation \textbf{GNN}), which consists of stacked residual local learning-based GCN layers and a attention-based propagation layer. 
Extracted node embedding, graph embedding from pre-trained auto-encoder and multi-frequency level encoding are concatenated to the original node features. 
Graph is firstly fed into stacked GCN layers.
For the forward process of attention-based propagation layer, we apply topological sorting on the circuit graph to calculate the order of each node and select nodes with the same order as a level.
We feed this single level into the attention-based propagation layer, and calculate AT for its vertices.
Note that we feed each level into attention-based propagation layer with ascending order, ensuring that when calculating AT of current level of nodes, AT of all their predecessors have already been calculated.
Nodes belonging to the same level are updated first, followed by nodes of next order until to the last.
This process realizes the asynchronous calculation of AT for each node, enabling the signal propagation along timing paths.
Our model predicts AT as a main task, with slew, net delay and cell delay prediction as auxiliary tasks following~\cite{guo2022timing}. 
Complete pipeline is shown in Fig.~\ref{fig:pipeline}. 

The total loss for second stage is as follows:
\begin{equation}
\label{eq:loss-training-level-GNN}
    {
    \small
    \begin{aligned}
         \mathcal{L}_{AS} & = \lVert M^{\psi_2}_{AS}(M^{\psi_1}(g, \mathbf{X}, E_\theta(g, \mathbf{X}))) - AS \rVert_2^2  \\
         \mathcal{L}_{CD} & = \lVert M^{\psi_3}_{CD}(M^{\psi_1}(g, \mathbf{X}, E_\theta(g, \mathbf{X}))) - CD \rVert_2^2  \\
         \mathcal{L}_{ND} & = \lVert M^{\psi_4}_{ND}(M^{\psi_1}(g, \mathbf{X}, E_\theta(g, \mathbf{X}))) - ND \rVert_2^2  \\
         \mathcal{L}_{GNN}& = \mathcal{L}_{AS} + \lambda_{CD} \mathcal{L}_{CD} + \lambda_{ND} \mathcal{L}_{ND},
    \end{aligned}
    }
\end{equation}
where $g$ is the circuit graph, $\psi_1$ to $\psi_4$ are trainable parameters of GNN, $\mathbf{X}$ is node feature. $CD, ND, AS$ are cell delay, net delay, AT concatenated with slew, $\lambda_{CD}, \lambda_{ND}$ are loss weights for cell delay and net delay.



\begin{table*}[!t]
    \centering
    
    {\fontsize{9pt}{10pt}\selectfont
    \begin{tabular}{ll|ccccccccccccccccccccc}
        \toprule

\multicolumn{2}{c|}{\multirow{2}{*}{Dataset}} & \multicolumn{6}{c}{$R^2_{uf}$ ($\uparrow$)}\\
\multicolumn{2}{c|}{} & baseline & w/o AE & w/o RLL & w/o MJA & w/o FT & PreRoutGNN\\
\midrule
\multirow{7}{*}{\rotatebox{90}{test}} & aes192 & 0.649 & 0.603 & 0.825 & 0.878 & \textbf{0.966} & \underline{0.946}\\
~ & jpeg\_encoder & 0.271 & \textbf{0.787} & -0.338 & 0.605 & 0.620 & \underline{0.774}\\
~ & spm & 0.696 & 0.883 & \underline{0.988} & 0.987 & 0.977 & \textbf{0.992}\\
~ & synth\_ram & 0.763 & 0.423 & \underline{0.975} & 0.958 & 0.949 & \textbf{0.984}\\
~ & usbf\_device & 0.317 & 0.659 & 0.676 & 0.844 & \underline{0.931} & \textbf{0.951}\\
~ & xtea & 0.781 & 0.921 & 0.976 & 0.955 & \underline{0.980} & \textbf{0.985}\\
~ & y\_huff & 0.552 & 0.766 & 0.883 & 0.885 & \textbf{0.934} & 0.901\\
\midrule
\multicolumn{2}{c|}{Avg. train} & 0.720 & 0.959 & 0.876 & 0.933 & \underline{0.963} & \textbf{0.964}\\
\multicolumn{2}{c|}{Avg. test} & 0.576 & 0.720 & 0.772 & 0.873 & \underline{0.908} & \textbf{0.933}\\

\bottomrule
\end{tabular}
}
\caption{
    Ablation studies. We remove one and only one module each time. 
    $R_{uf}^2$ of slack prediction is reported. 
    AE: global pre-trained auto-encoder, RLL: residual local learning, MJA: multi-head joint attention, FT: fine-tune the pre-trained encoder.
    {Bold}/{underline} indicates best/second.
    }
\label{tab:ablation-study-1}
\end{table*}

\section*{Experiments}
\subsection*{Evaluation Protocols}

\textbf{Benchmarks and compared methods.}
We evaluate our approach on public benchmark~\cite{guo2022timing} with 21 real world circuits.
Given the circuit after standard cell placement, the objective is to predict slack, slew, cell delay and net delay after routing. 
The compared methods include GCNII~\cite{chen2020simple}, GAT~\cite{velivckovic2018graph}, GINE~\cite{hu2019strategies} and TimingGCN~\cite{guo2022timing}.

\textbf{Metrics.}
We evaluate the prediction of slack with the $R^2$ determination coefficient score as follows: 
\begin{equation}
\label{eq:R2}
\scalebox{1.0}{$
R^2 = 1 - \frac{\frac{1}{n}\sum_{i=1}^{n} (y_i - \hat{y}_i)^2}{\frac{1}{n}\sum_{i=1}^{n} (y_i - \bar{y})^2} = 1- \frac{\mathrm{MSE(\mathbf{y, \hat{y}})}}{\mathrm{VAR(\mathbf{y})}},
$}
\end{equation}
where $y_i$ is ground truth, $\hat{y}$ is prediction and $\bar{y}$ is the mean value of ground truth. Since slack has four channels (EL/RF), we can calculate $R^2$ in two ways: \textbf{1) \emph{un-flatten}} $R_{uf}^2$ and  \textbf{ 2) \emph{flatten}} $R_f^2$ .  
In $R_{uf}^2$, we calculate $R^2$ on \emph{un-flatten} data with shape $\mathbb{R}^{n \times 4}$, where $n$ is the number of nodes. 
We calculate $R^2$ on each channel independently and average these four $R^2$ as the final $R_{uf}^2$. 
In $R_f^2$, we \emph{flatten} the data to $ \mathbb{R}^{4n}$, merging four different channels together and the number of nodes increases from $n$ to $4n$. 
Under $R_f^2$, previous method achieves 0.866.
However, under $R_{uf}^2$, it only achieves 0.576.

\begin{table*}[!tb]
    \centering
    
    {\fontsize{9pt}{10pt}\selectfont
    \begin{tabular}{ll|cc|cc|ccccccccccc}
        \toprule

\multicolumn{2}{c|}{\multirow{2}{*}{Dataset}} & \multicolumn{2}{c|}{$R^2_{uf}$ slew ($\uparrow$)} & \multicolumn{2}{c|}{$R^2_{uf}$ net delay ($\uparrow$)} & \multicolumn{2}{c}{$R^2_{uf}$ cell delay ($\uparrow$)}\\
\multicolumn{2}{c|}{} & TimingGCN & PreRoutGNN & TimingGCN & PreRoutGNN & TimingGCN & PreRoutGNN\\
\midrule
\multirow{7}{*}{\rotatebox{90}{test}} & aes192 & 0.816 $\pm$ 0.037 & \textbf{0.967 $\pm$ 0.005} & 0.954 $\pm$ 0.010 & \textbf{0.979 $\pm$ 0.001} & 0.702 $\pm$ 0.244 & \textbf{0.975 $\pm$ 0.005}\\
~ & jpeg\_encoder & 0.783 $\pm$ 0.038 & \textbf{0.949 $\pm$ 0.000} & 0.966 $\pm$ 0.004 & \textbf{0.981 $\pm$ 0.001} & 0.636 $\pm$ 0.233 & \textbf{0.947 $\pm$ 0.011}\\
~ & spm & 0.904 $\pm$ 0.017 & \textbf{0.981 $\pm$ 0.001} & 0.920 $\pm$ 0.010 & \textbf{0.940 $\pm$ 0.019} & 0.831 $\pm$ 0.037 & \textbf{0.981 $\pm$ 0.003}\\
~ & synth\_ram & 0.764 $\pm$ 0.042 & \textbf{0.846 $\pm$ 0.073} & 0.929 $\pm$ 0.014 & \textbf{0.974 $\pm$ 0.001} & 0.533 $\pm$ 0.332 & \textbf{0.682 $\pm$ 0.249}\\
~ & usbf\_device & 0.859 $\pm$ 0.023 & \textbf{0.924 $\pm$ 0.005} & 0.968 $\pm$ 0.001 & \textbf{0.970 $\pm$ 0.003} & 0.824 $\pm$ 0.072 & \textbf{0.976 $\pm$ 0.004}\\
~ & xtea & 0.824 $\pm$ 0.072 & \textbf{0.978 $\pm$ 0.002} & 0.944 $\pm$ 0.011 & \textbf{0.963 $\pm$ 0.000} & 0.574 $\pm$ 0.303 & \textbf{0.968 $\pm$ 0.007}\\
~ & y\_huff & 0.880 $\pm$ 0.026 & \textbf{0.896 $\pm$ 0.002} & 0.940 $\pm$ 0.014 & \textbf{0.964 $\pm$ 0.005} & \textbf{0.844 $\pm$ 0.066} & 0.813 $\pm$ 0.118\\
\midrule
\multicolumn{2}{c|}{Avg. train} & 0.883 & \textbf{0.968} & 0.982 & \textbf{0.992} & 0.832 & \textbf{0.980}\\
\multicolumn{2}{c|}{Avg. test} & 0.833 & \textbf{0.934} & 0.946 & \textbf{0.967} & 0.706 & \textbf{0.906}\\

\bottomrule
\end{tabular}
}
\caption{Comparison of other tasks, including prediction of slew, net delay and cell delay. 
}

\label{tab:auxiliary-tasks}
\end{table*}

After reforming $R^2$ in Eq.~\ref{eq:R2}, we calculate it with VAR (variance) of $\mathbf{y}$, MSE(Mean Squared Error) between $\mathbf{y}$ and $\mathbf{\hat{y}}$.
After merging four different channels into a single channel, both MSE and VAR increase, but VAR increases more than MSE, resulting in the improvement of $R^2$. 
Let $k = \mathrm{VAR(\mathbf{y})} / \mathrm{MSE(\mathbf{y, \hat{y}})}$. 
Since slack has four channels, we compute $k$ on each channel and average them to get the mean $k_{uf}$. 
Flatten ratio $k_f$ is computed in the same way, except for flattening slack from $\mathbb{R}^{n \times 4}$ to $\mathbb{R}^{4n}$ with only one channel. 
With $k_f / k_{uf} > 1$ we can see that, after flattening the multi-channel slack, VAR increases faster and higher than MSE, leading to the improvement about 0.3 of $R^2$.

We think the `distance', e.g. Kullback-Leibler divergence, between distributions under different corners is high. 
The seemingly high flatten $R^2_f$ cannot reflect the real performance of model, and it is an inappropriate evaluation metric.
Thus, it is more reasonable to calculate $R^2$ on each channel independently and use $R_{uf}^2$ as the final evaluation metric.

\subsection*{Results and Further Studies}
\textbf{Overall performance comparison.} 
We use $R_{uf}^2$ as primary metric. 
In Table~\ref{tab:main-results-random-seeds}, our approach achieves $R_{uf}^2$ of 0.93, while the second best method only achieves 0.59. 

\textbf{Evaluation on other tasks.}
We also compare effectiveness on other tasks, including prediction of slew, net delay and cell delay.
In Table~\ref{tab:auxiliary-tasks}, for $R_{uf}^2$ on testing set, we achieve 0.93, 0.97, 0.91, surpassing the best baseline achieving 0.83, 0.95, 0.71 respectively, demonstrating that our model is also adaptive to multiple pre-routing timing prediction tasks.

\begin{table}[!tb]
    \centering
    
    {\fontsize{9pt}{10pt}\selectfont
    \begin{tabular}{ll|ccccccccccccccc}
        \toprule

\multicolumn{2}{c|}{\multirow{2}{*}{Dataset}} & \multicolumn{2}{c}{$R^2_{uf}$ ($\uparrow$)}\\
\multicolumn{2}{c|}{} & TimingGCN & + Pre-training\\
\midrule
\multirow{7}{*}{\rotatebox{90}{test}} & aes192 & 0.384 $\pm$ 0.205 & \textbf{0.654 $\pm$ 0.097}\\
~ & jpeg\_encoder & 0.478 $\pm$ 0.131 & \textbf{0.505 $\pm$ 0.268}\\
~ & spm & 0.802 $\pm$ 0.083 & \textbf{0.944 $\pm$ 0.020}\\
~ & synth\_ram & 0.735 $\pm$ 0.079 & \textbf{0.943 $\pm$ 0.023}\\
~ & usbf\_device & 0.347 $\pm$ 0.119 & \textbf{0.525 $\pm$ 0.184}\\
~ & xtea & 0.888 $\pm$ 0.062 & \textbf{0.946 $\pm$ 0.026}\\
~ & y\_huff & 0.490 $\pm$ 0.146 & \textbf{0.816 $\pm$ 0.046}\\
\midrule
\multicolumn{2}{c|}{Avg. train} & 0.733  & \textbf{0.781}\\
\multicolumn{2}{c|}{Avg. test} & 0.589  & \textbf{0.762}\\

\bottomrule
\end{tabular}
}
\caption{
    Slack prediction with $R_{uf}^2$. 
    After applying pre-trained graph encoder into TimingGCN, $R_{uf}^2$ improves 29 \%, showing that our pre-trained graph encoder can serve as a plug-and-play module for training. 
    }
\label{tab:TimingGCN-pre-training}
\end{table}

\textbf{Ablation studies.}
We ablate the effectiveness of each module of our method. 
In Table~\ref{tab:ablation-study-1}, we remove one and only one component each time.
After removing global pre-training or RLL, $R_{uf}^2$ drops dramatically, revealing the importance of global view and local modeling for signal delay.
We also compare fine-tuning with freezing the pre-trained encoder.
Even without fine-tuning, our model behaves well, only with a small drop 0.02 of $R_{uf}^2$, suggesting the importance of  global circuit pre-training: providing a global view and a representation of circuit.
We can also apply our  global circuit pre-trained graph encoder to other baselines.
Training other baselines along with fine-tuning graph encoder achieves better results with improvement of 29\% in Table~\ref{tab:TimingGCN-pre-training}.
The pre-trained graph encoder serves as a plug-and-play module for downstream tasks.

\begin{figure}[!tb]
    \centering
    \includegraphics[width=0.9\linewidth]{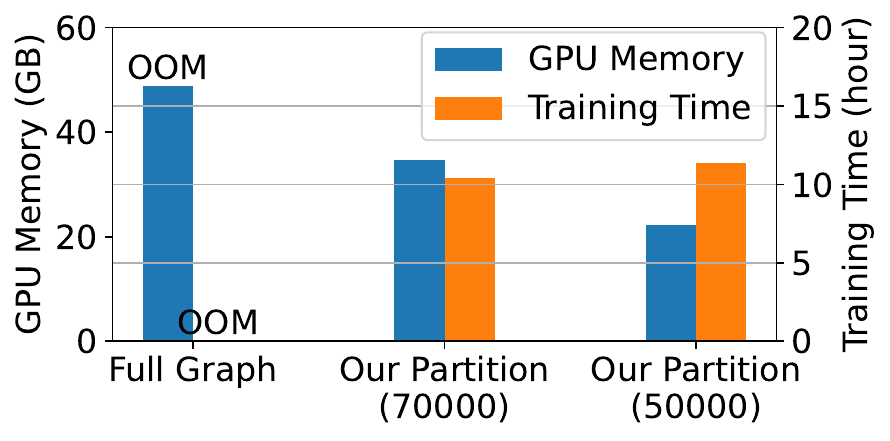}
    \caption{
    GPU memory cost and training time comparison between training on whole graphs and partitioned graphs. 
    Number in brackets indicates the maximum sub-graph size.
    }
    \label{fig:GPU_memory-runtime}
\end{figure}

\textbf{Runtime and GPU Memory cost.}
In Fig.~\ref{fig:GPU_memory-runtime}, our method significantly reduces the peak GPU memory cost. Empirically, maximum sub-graph size 50,000 performs well.
For large-scale circuits with about 300,000 pins, inference can be finished within one second.

\section*{Conclusion}
We propose a novel pre-routing timing prediction method, PreRoutGNN, to address the signal decay and error accumulation issues.
We use global circuit  pre-training to provide global view, and devise residual local learning, MJA for long timing path modeling.
Besides, our global circuit pre-training can serve as a plug-and-play module for various timing prediction GNN.
For tractability on large-scale circuits, we also devise order preserving partition scheme to reduce memory cost while preserving the topological dependencies.
We believe our works demonstrate accurate modeling of signal propagation along timing paths.

\section*{Acknowledgements}
This work was partly supported by China Key Research and Development Program (2020AAA0107600) and NSFC (92370132, 62222607).

\clearpage
\bibliography{aaai24}

\begin{thebibliography}{49}
\providecommand{\natexlab}[1]{#1}

\bibitem[{Ajayi et~al.(2019)Ajayi, Blaauw, Chan, Cheng, Chhabria, Choo, Coltella, Dobre, Dreslinski, Foga{\c{c}}a et~al.}]{ajayi2019openroad}
Ajayi, T.; Blaauw, D.; Chan, T.; Cheng, C.; Chhabria, V.; Choo, D.; Coltella, M.; Dobre, S.; Dreslinski, R.; Foga{\c{c}}a, M.; et~al. 2019.
\newblock OpenROAD: Toward a Self-Driving, Open-Source Digital Layout Implementation Tool Chain.
\newblock \emph{Proc. GOMACTECH}.

\bibitem[{Bao et~al.(2022)Bao, Dong, Piao, and Wei}]{bao2022beit}
Bao, H.; Dong, L.; Piao, S.; and Wei, F. 2022.
\newblock Beit: Bert pre-training of image transformers.
\newblock In \emph{ICLR}.

\bibitem[{Barboza et~al.(2019)Barboza, Shukla, Chen, and Hu}]{barboza2019machine}
Barboza, E.~C.; Shukla, N.; Chen, Y.; and Hu, J. 2019.
\newblock Machine learning-based pre-routing timing prediction with reduced pessimism.
\newblock In \emph{DAC}.

\bibitem[{Brown et~al.(2020)Brown, Mann, Ryder, Subbiah, Kaplan, Dhariwal, Neelakantan, Shyam, Sastry, Askell et~al.}]{brown2020language}
Brown, T.; Mann, B.; Ryder, N.; Subbiah, M.; Kaplan, J.~D.; Dhariwal, P.; Neelakantan, A.; Shyam, P.; Sastry, G.; Askell, A.; et~al. 2020.
\newblock Language models are few-shot learners.
\newblock \emph{NeurIPS}.

\bibitem[{Chen et~al.(2020)Chen, Wei, Huang, Ding, and Li}]{chen2020simple}
Chen, M.; Wei, Z.; Huang, Z.; Ding, B.; and Li, Y. 2020.
\newblock Simple and deep graph convolutional networks.
\newblock In \emph{ICML}.

\bibitem[{Cheng et~al.(2022)Cheng, Lyu, Li, Ye, Hao, and Yan}]{cheng2022policy}
Cheng, R.; Lyu, X.; Li, Y.; Ye, J.; Hao, J.; and Yan, J. 2022.
\newblock The Policy-gradient Placement and Generative Routing Neural Networks for Chip Design.
\newblock In \emph{NeurIPS}.

\bibitem[{Cheng and Yan(2021)}]{cheng2021joint}
Cheng, R.; and Yan, J. 2021.
\newblock On joint learning for solving placement and routing in chip design.
\newblock In \emph{NeurIPS}.

\bibitem[{Chowdhury et~al.(2022)Chowdhury, Tan, Carey, Jain, Karri, and Garg}]{chowdhury2022too}
Chowdhury, A.~B.; Tan, B.; Carey, R.; Jain, T.; Karri, R.; and Garg, S. 2022.
\newblock Too Big to Fail? Active Few-Shot Learning Guided Logic Synthesis.
\newblock \emph{arXiv preprint arXiv:2204.02368}.

\bibitem[{Defferrard, Bresson, and Vandergheynst(2016)}]{defferrard2016convolutional}
Defferrard, M.; Bresson, X.; and Vandergheynst, P. 2016.
\newblock Convolutional neural networks on graphs with fast localized spectral filtering.
\newblock In \emph{NeurIPS}.

\bibitem[{Devlin et~al.(2018)Devlin, Chang, Lee, and Toutanova}]{devlin2018bert}
Devlin, J.; Chang, M.-W.; Lee, K.; and Toutanova, K. 2018.
\newblock Bert: Pre-training of deep bidirectional transformers for language understanding.
\newblock \emph{arXiv preprint arXiv:1810.04805}.

\bibitem[{Du et~al.(2023)Du, Wang, Zhong, and Yan}]{du2023hubrouter}
Du, X.; Wang, C.; Zhong, R.; and Yan, J. 2023.
\newblock Hubrouter: Learning global routing via hub generation and pin-hub connection.
\newblock In \emph{NeurIPS}.

\bibitem[{Ghose et~al.(2021)Ghose, Zhang, Zhang, Li, Liu, and Coates}]{ghose2021generalizable}
Ghose, A.; Zhang, V.; Zhang, Y.; Li, D.; Liu, W.; and Coates, M. 2021.
\newblock Generalizable Cross-Graph Embedding for GNN-based Congestion Prediction.
\newblock In \emph{ICCAD}.

\bibitem[{Guo et~al.(2022)Guo, Liu, Gu, Zhang, Pan, and Lin}]{guo2022timing}
Guo, Z.; Liu, M.; Gu, J.; Zhang, S.; Pan, D.~Z.; and Lin, Y. 2022.
\newblock A timing engine inspired graph neural network model for pre-routing slack prediction.
\newblock In \emph{DAC}.

\bibitem[{Hartigan and Wong(1979)}]{hartigan1979algorithm}
Hartigan, J.~A.; and Wong, M.~A. 1979.
\newblock Algorithm AS 136: A k-means clustering algorithm.
\newblock \emph{Journal of the royal statistical society. series c (applied statistics)}.

\bibitem[{Hu, Schaeffer, and Garg(2015)}]{hu2015tau}
Hu, J.; Schaeffer, G.; and Garg, V. 2015.
\newblock TAU 2015 contest on incremental timing analysis.
\newblock In \emph{ICCAD}.

\bibitem[{Hu, Sinha, and Keller(2014)}]{hu2014tau}
Hu, J.; Sinha, D.; and Keller, I. 2014.
\newblock TAU 2014 contest on removing common path pessimism during timing analysis.
\newblock In \emph{ISPD}.

\bibitem[{Hu et~al.(2020)Hu, Liu, Gomes, Zitnik, Liang, Pande, and Leskovec}]{hu2019strategies}
Hu, W.; Liu, B.; Gomes, J.; Zitnik, M.; Liang, P.; Pande, V.; and Leskovec, J. 2020.
\newblock Strategies for pre-training graph neural networks.
\newblock In \emph{ICLR}.

\bibitem[{Karypis and Kumar(1998)}]{karypis1998software}
Karypis, G.; and Kumar, V. 1998.
\newblock A software package for partitioning unstructured graphs, partitioning meshes, and computing fill-reducing orderings of sparse matrices.
\newblock \emph{University of Minnesota, Department of Computer Science and Engineering, Army HPC Research Center, Minneapolis, MN}.

\bibitem[{Khan et~al.(2023)Khan, Shi, Li, and Xu}]{khan2023deepseq}
Khan, S.; Shi, Z.; Li, M.; and Xu, Q. 2023.
\newblock DeepSeq: Deep Sequential Circuit Learning.
\newblock \emph{arXiv preprint arXiv:2302.13608}.

\bibitem[{Kingma and Ba(2014)}]{kingma2014adam}
Kingma, D.~P.; and Ba, J. 2014.
\newblock Adam: A method for stochastic optimization.
\newblock \emph{arXiv preprint arXiv:1412.6980}.

\bibitem[{Kipf and Welling(2017)}]{kipf2016semi}
Kipf, T.~N.; and Welling, M. 2017.
\newblock Semi-supervised classification with graph convolutional networks.
\newblock In \emph{ICLR}.

\bibitem[{Lai et~al.(2023)Lai, Liu, Tang, Wang, Hao, and Luo}]{lai2023chipformer}
Lai, Y.; Liu, J.; Tang, Z.; Wang, B.; Hao, J.; and Luo, P. 2023.
\newblock Chipformer: Transferable chip placement via offline decision transformer.
\newblock In \emph{ICML}.

\bibitem[{Lai, Mu, and Luo(2022)}]{lai2022maskplace}
Lai, Y.; Mu, Y.; and Luo, P. 2022.
\newblock Maskplace: Fast chip placement via reinforced visual representation learning.
\newblock \emph{NeurIPS}.

\bibitem[{Lasser(1961)}]{lasser1961topological}
Lasser, D.~J. 1961.
\newblock Topological ordering of a list of randomly-numbered elements of a network.
\newblock \emph{Communications of the ACM}.

\bibitem[{Li et~al.(2022)Li, Khan, Shi, Wang, Yu, and Xu}]{li2022deepgate}
Li, M.; Khan, S.; Shi, Z.; Wang, N.; Yu, H.; and Xu, Q. 2022.
\newblock Deepgate: Learning neural representations of logic gates.
\newblock In \emph{DAC}.

\bibitem[{Liao et~al.(2022)Liao, Liu, Chen, Lv, Lin, and Yu}]{liao2022dreamplace}
Liao, P.; Liu, S.; Chen, Z.; Lv, W.; Lin, Y.; and Yu, B. 2022.
\newblock DREAMPlace 4.0: Timing-driven global placement with momentum-based net weighting.
\newblock In \emph{DATE}.

\bibitem[{Lin et~al.(2019)Lin, Dhar, Li, Ren, Khailany, and Pan}]{lin2019dreamplace}
Lin, Y.; Dhar, S.; Li, W.; Ren, H.; Khailany, B.; and Pan, D.~Z. 2019.
\newblock Dreamplace: Deep learning toolkit-enabled gpu acceleration for modern vlsi placement.
\newblock In \emph{DAC}.

\bibitem[{Lu et~al.(2015)Lu, Chen, Chang, Sha, Huang, Teng, and Cheng}]{lu2015eplace}
Lu, J.; Chen, P.; Chang, C.-C.; Sha, L.; Huang, D. J.-H.; Teng, C.-C.; and Cheng, C.-K. 2015.
\newblock ePlace: Electrostatics-based placement using fast fourier transform and Nesterov's method.
\newblock \emph{ACM Transactions on Design Automation of Electronic Systems (TODAES)}.

\bibitem[{Mildenhall et~al.(2021)Mildenhall, Srinivasan, Tancik, Barron, Ramamoorthi, and Ng}]{mildenhall2021nerf}
Mildenhall, B.; Srinivasan, P.~P.; Tancik, M.; Barron, J.~T.; Ramamoorthi, R.; and Ng, R. 2021.
\newblock Nerf: Representing scenes as neural radiance fields for view synthesis.
\newblock \emph{Communications of the ACM}.

\bibitem[{Mirhoseini et~al.(2021)Mirhoseini, Goldie, Yazgan, Jiang, Songhori, Wang, Lee, Johnson, Pathak, Nazi et~al.}]{mirhoseini2021graph}
Mirhoseini, A.; Goldie, A.; Yazgan, M.; Jiang, J.~W.; Songhori, E.; Wang, S.; Lee, Y.-J.; Johnson, E.; Pathak, O.; Nazi, A.; et~al. 2021.
\newblock A graph placement methodology for fast chip design.
\newblock \emph{Nature}.

\bibitem[{Park and Kim(2022)}]{park2022vision}
Park, N.; and Kim, S. 2022.
\newblock How do vision transformers work?
\newblock In \emph{ICLR}.

\bibitem[{Paszke et~al.(2019)Paszke, Gross, Massa, Lerer, Bradbury, Chanan, Killeen, Lin, Gimelshein, Antiga et~al.}]{paszke2019pytorch}
Paszke, A.; Gross, S.; Massa, F.; Lerer, A.; Bradbury, J.; Chanan, G.; Killeen, T.; Lin, Z.; Gimelshein, N.; Antiga, L.; et~al. 2019.
\newblock Pytorch: An imperative style, high-performance deep learning library.
\newblock In \emph{NeurIPS}.

\bibitem[{Sanders and Schulz(2013)}]{sanders2013think}
Sanders, P.; and Schulz, C. 2013.
\newblock Think locally, act globally: Highly balanced graph partitioning.
\newblock In \emph{Experimental Algorithms: 12th International Symposium, SEA 2013, Rome, Italy, June 5-7, 2013. Proceedings 12}.

\bibitem[{Vaswani et~al.(2017)Vaswani, Shazeer, Parmar, Uszkoreit, Jones, Gomez, Kaiser, and Polosukhin}]{vaswani2017attention}
Vaswani, A.; Shazeer, N.; Parmar, N.; Uszkoreit, J.; Jones, L.; Gomez, A.~N.; Kaiser, {\L}.; and Polosukhin, I. 2017.
\newblock Attention is all you need.
\newblock In \emph{NeurIPS}.

\bibitem[{Veli{\v{c}}kovi{\'c} et~al.(2018)Veli{\v{c}}kovi{\'c}, Cucurull, Casanova, Romero, Lio, and Bengio}]{velivckovic2018graph}
Veli{\v{c}}kovi{\'c}, P.; Cucurull, G.; Casanova, A.; Romero, A.; Lio, P.; and Bengio, Y. 2018.
\newblock Graph attention networks.
\newblock In \emph{ICLR}.

\bibitem[{Wang et~al.(2022{\natexlab{a}})Wang, Shen, Li, Hao, Liu, Huang, Wu, Lin, Chen, and Heng}]{wang2022lhnn}
Wang, B.; Shen, G.; Li, D.; Hao, J.; Liu, W.; Huang, Y.; Wu, H.; Lin, Y.; Chen, G.; and Heng, P.~A. 2022{\natexlab{a}}.
\newblock LHNN: Lattice hypergraph neural network for VLSI congestion prediction.
\newblock In \emph{DAC}.

\bibitem[{Wang et~al.(2019)Wang, Zheng, Ye, Gan, Li, Song, Zhou, Ma, Yu, Gai et~al.}]{wang2019deep}
Wang, M.; Zheng, D.; Ye, Z.; Gan, Q.; Li, M.; Song, X.; Zhou, J.; Ma, C.; Yu, L.; Gai, Y.; et~al. 2019.
\newblock Deep graph library: A graph-centric, highly-performant package for graph neural networks.
\newblock \emph{arXiv preprint arXiv:1909.01315}.

\bibitem[{Wang et~al.(2022{\natexlab{b}})Wang, Zheng, Chen, and Wang}]{wang2022anti}
Wang, P.; Zheng, W.; Chen, T.; and Wang, Z. 2022{\natexlab{b}}.
\newblock Anti-oversmoothing in deep vision transformers via the fourier domain analysis: From theory to practice.
\newblock In \emph{ICLR}.

\bibitem[{Wu et~al.(2023{\natexlab{a}})Wu, Yang, Zhao, He, Wipf, and Yan}]{wu2023difformer}
Wu, Q.; Yang, C.; Zhao, W.; He, Y.; Wipf, D.; and Yan, J. 2023{\natexlab{a}}.
\newblock Difformer: Scalable (graph) transformers induced by energy constrained diffusion.
\newblock In \emph{ICLR}.

\bibitem[{Wu et~al.(2022)Wu, Zhao, Li, Wipf, and Yan}]{wu2022nodeformer}
Wu, Q.; Zhao, W.; Li, Z.; Wipf, D.~P.; and Yan, J. 2022.
\newblock Nodeformer: A scalable graph structure learning transformer for node classification.
\newblock In \emph{NeurIPS}.

\bibitem[{Wu et~al.(2023{\natexlab{b}})Wu, Zhao, Yang, Zhang, Nie, Jiang, Bian, and Yan}]{wu2023simplifying}
Wu, Q.; Zhao, W.; Yang, C.; Zhang, H.; Nie, F.; Jiang, H.; Bian, Y.; and Yan, J. 2023{\natexlab{b}}.
\newblock Simplifying and Empowering Transformers for Large-Graph Representations.
\newblock In \emph{NeurIPS}.

\bibitem[{Xie et~al.(2018)Xie, Huang, Fang, Ren, Fang, Chen, and Hu}]{xie2018routenet}
Xie, Z.; Huang, Y.-H.; Fang, G.-Q.; Ren, H.; Fang, S.-Y.; Chen, Y.; and Hu, J. 2018.
\newblock RouteNet: Routability prediction for mixed-size designs using convolutional neural network.
\newblock In \emph{ICCAD}.

\bibitem[{Xie et~al.(2021)Xie, Liang, Xu, Hu, Duan, and Chen}]{xie2021net2}
Xie, Z.; Liang, R.; Xu, X.; Hu, J.; Duan, Y.; and Chen, Y. 2021.
\newblock Net2: A graph attention network method customized for pre-placement net length estimation.
\newblock In \emph{ASP-DAC}.

\bibitem[{Xu et~al.(2019)Xu, Hu, Leskovec, and Jegelka}]{xu2019powerful}
Xu, K.; Hu, W.; Leskovec, J.; and Jegelka, S. 2019.
\newblock How powerful are graph neural networks?
\newblock In \emph{ICLR}.

\bibitem[{Yang, He, and Cao(2022)}]{yang2022pre}
Yang, T.; He, G.; and Cao, P. 2022.
\newblock Pre-routing path delay estimation based on transformer and residual framework.
\newblock In \emph{ASP-DAC}.

\bibitem[{Yang et~al.(2022)Yang, Li, Zhang, Zhang, Song, Hao et~al.}]{yang2022versatile}
Yang, Z.; Li, D.; Zhang, Y.; Zhang, Z.; Song, G.; Hao, J.; et~al. 2022.
\newblock Versatile Multi-stage Graph Neural Network for Circuit Representation.
\newblock In \emph{NeurIPS}.

\bibitem[{Yuan et~al.(2023)Yuan, Wang, Ye, Yuan, Hao, and Yan}]{EasySO}
Yuan, J.; Wang, P.; Ye, J.; Yuan, M.; Hao, J.; and Yan, J. 2023.
\newblock EasySO: Exploration-enhanced Reinforcement Learning for Logic Synthesis Sequence Optimization and a Comprehensive RL Environment.
\newblock In \emph{ICCAD}.

\bibitem[{Zhao et~al.(2023)Zhao, Wu, Yang, and Yan}]{zhao2023graphglow}
Zhao, W.; Wu, Q.; Yang, C.; and Yan, J. 2023.
\newblock GraphGLOW: Universal and Generalizable Structure Learning for Graph Neural Networks.
\newblock In \emph{SIGKDD}.

\bibitem[{Zhou et~al.(2022)Zhou, Ye, Pui, Shao, Zhang, Wang, Hao, Chen, and Heng}]{zhou2022heterogeneous}
Zhou, X.; Ye, J.; Pui, C.-W.; Shao, K.; Zhang, G.; Wang, B.; Hao, J.; Chen, G.; and Heng, P.~A. 2022.
\newblock Heterogeneous Graph Neural Network-based Imitation Learning for Gate Sizing Acceleration.
\newblock In \emph{ICCAD}.

\end{thebibliography}

\clearpage
\appendix
\section{Details about Datasets}
Datasets we use contain 21 real-world OpenROAD~\cite{ajayi2019openroad} circuits.
In our experiments, the datasets are split into training and testing sets in line with the same settings in ~\cite{guo2022timing}. 
Details of dataset are shown in Table~\ref{tab:dataset-information}.
\begin{table}[!b]
    \centering
    \adjustbox{width=1.0\linewidth}{
    \begin{tabular}{ll|cccccccccccccc}
        \toprule
         \multicolumn{2}{c|}{\multirow{2}{*}{Dataset}} & \multirow{2}{*}{\# Node} & \multicolumn{3}{c}{\# Edge} & \multirow{2}{*}{\# Level} \\
         \multicolumn{2}{c|}{} & ~ & cell & net & net\_inv & ~ \\
\midrule
\multirow{14}{*}{train} & BM64 & 38458 & 25334 & 27843 & 27843 & 100\\
~ & aes128 & 211045 & 138457 & 148997 & 148997 & 126\\
~ & aes256 & 290955 & 189262 & 207414 & 207414 & 124\\
~ & aes\_cipher & 59777 & 41411 & 42671 & 42671 & 70\\
~ & blabla & 55568 & 35689 & 39853 & 39853 & 154\\
~ & cic\_decimator & 3131 & 2102 & 2232 & 2232 & 40\\
~ & des & 60541 & 41845 & 44478 & 44478 & 48\\
~ & genericfir & 38827 & 25013 & 28845 & 28845 & 32\\
~ & picorv32a & 58676 & 40208 & 43047 & 43047 & 168\\
~ & salsa20 & 78486 & 52895 & 57737 & 57737 & 160\\
~ & usb & 3361 & 2189 & 2406 & 2406 & 34\\
~ & usb\_cdc\_core & 7406 & 4869 & 5200 & 5200 & 74\\
~ & wbqspiflash & 9672 & 6454 & 6798 & 6798 & 112\\
~ & zipdiv & 4398 & 2913 & 3102 & 3102 & 72\\
\midrule
\multirow{7}{*}{test} & aes192 & 234211 & 152910 & 165350 & 165350 & 134\\
~ & jpeg\_encoder & 238216 & 167960 & 176737 & 176737 & 124\\
~ & spm & 1121 & 700 & 765 & 765 & 18\\
~ & synth\_ram & 25910 & 16782 & 19024 & 19024 & 24\\
~ & usbf\_device & 66345 & 42226 & 46241 & 46241 & 110\\
~ & xtea & 10213 & 6882 & 7151 & 7151 & 114\\
~ & y\_huff & 48216 & 30612 & 33689 & 33689 & 58\\

\bottomrule
\end{tabular}
}
    \caption{Dataset information: number of nodes, edges and levels for datasets.}
\label{tab:dataset-information}
\end{table}

\section{Implementation Details}
We implement our model and algorithm with PyTorch~\cite{paszke2019pytorch} and  Deep Graph Library (DGL)~\cite{wang2019deep} framework. 
We train and test our model on a Linux server with one NVIDIA RTX 3090 GPU with 24 GB CUDA memory, 64 AMD Ryzen Threadripper 3970X 32-Core Processor at 2.20 GHz and 128 GB RAM. 
We use Adam~\cite{kingma2014adam} optimizer with learning rate $lr=5 \times 10^{-4}, \beta_1 = 0.9, \beta_2 = 0.999$. 
For all models, we train them with 20000 epochs.

\section{Computation of Slack}
In circuit, electrical signal travels from primary inputs (input ports) to endpoints or primary outputs (output ports) along timing paths.
Since signal cannot travel with an infinite speed, there are delay in circuit, and the instant when signal arrives at a pin is defined as \emph{Arrival Time (AT)}.
Considering the timing constraints in circuits, e.g. the instant when signal arrives a pin cannot be too late, there are also \emph{Required Arrival Time (RAT)} for each pin, which means that the instant when signal arrives at this pin cannot be later than RAT.

Based on AT and RAT, the slack is defined as the difference between RAT and AT  as follows:
\begin{equation}
    \label{eq:slack-appendix}
    \begin{aligned}
        slack^E & = AT^E - RAT^E\\
        slack^L & = RAT^L - AT^L,
    \end{aligned}
\end{equation}
where $E,L$ means early and late respectively, and a positive slack means that the timing constraint is satisfied. 

\section{Experiments}

\subsection{Flatten $R_f^2$ and Un-Flatten $R_{uf}^2$}
Under $R_f^2$, previous method achieves 0.866.
However, under $R_{uf}^2$, it only achieves 0.576 as shown in Table~\ref{tab:flatten-or-not}.
With $k_f / k_{uf} > 1$ in Table~\ref{tab:flatten-or-not} 
we can see that, after flattening the multi-channel slack, VAR increases faster and higher than MSE, leading to the improvement about 0.3 of $R^2$. 
\begin{table}[!tb]
    \centering

    \adjustbox{width=1.0\linewidth}{
        \begin{tabular}{ll|cccccccccccccc}
             \toprule
             \multicolumn{2}{c|}{TimingGCN} & $k_{uf}$ & $k_{f}$ & $k_f / k_{uf}$ & $R^2_{uf}$ & $R^2_f$\\

 \midrule
\multirow{14}{*}{\rotatebox{90}{train}} & BM64 & 4.322 & 11.769 & \textbf{2.723} & 0.695 & \textbf{0.915}\\
~ & aes128 & 3.867 & 23.190 & \textbf{5.997} & -0.201 & \textbf{0.957}\\
~ & aes256 & 6.753 & 29.780 & \textbf{4.410} & 0.649 & \textbf{0.966}\\
~ & aes\_cipher & 5.877 & 36.984 & \textbf{6.293} & 0.614 & \textbf{0.973}\\
~ & blabla & 64.797 & 69.614 & \textbf{1.074} & 0.980 & \textbf{0.986}\\
~ & cic\_decimator & 56.084 & 81.282 & \textbf{1.449} & 0.979 & \textbf{0.988}\\
~ & des & 28.218 & 98.766 & \textbf{3.500} & 0.894 & \textbf{0.990}\\
~ & genericfir & 1.814 & 7.154 & \textbf{3.944} & 0.142 & \textbf{0.860}\\
~ & picorv32a & 5.860 & 16.753 & \textbf{2.859} & 0.811 & \textbf{0.940}\\
~ & salsa20 & 6.989 & 12.208 & \textbf{1.747} & 0.648 & \textbf{0.918}\\
~ & usb & 76.626 & 62.343 & 0.814 & 0.971 & \textbf{0.984}\\
~ & usb\_cdc\_core & 111.328 & 39.247 & 0.353 & 0.971 & \textbf{0.975}\\
~ & wbqspiflash & 52.909 & 50.267 & 0.950 & 0.972 & \textbf{0.980}\\
~ & zipdiv & 92.350 & 21.805 & 0.236 & \textbf{0.958} & 0.954\\
\midrule
\multirow{7}{*}{\rotatebox{90}{test}} & aes192 & 4.798 & 25.162 & \textbf{5.244} & 0.649 & \textbf{0.960}\\
~ & jpeg\_encoder & 1.717 & 10.627 & \textbf{6.190} & 0.271 & \textbf{0.906}\\
~ & spm & 4.931 & 4.189 & 0.850 & 0.696 & \textbf{0.761}\\
~ & synth\_ram & 7.898 & 5.301 & 0.671 & 0.763 & \textbf{0.811}\\
~ & usbf\_device & 2.723 & 12.740 & \textbf{4.678} & 0.317 & \textbf{0.922}\\
~ & xtea & 11.444 & 4.211 & 0.368 & \textbf{0.781} & 0.763\\
~ & y\_huff & 6.311 & 17.172 & \textbf{2.721} & 0.552 & \textbf{0.942}\\
\midrule
\multicolumn{2}{c|}{Avg. train} & 36.985 & 40.083 & \textbf{2.596} & 0.720 & \textbf{0.956}\\
\multicolumn{2}{c|}{Avg. test} & 5.689 & 11.343 & \textbf{2.960} & 0.576 & \textbf{0.866}\\

\bottomrule
\end{tabular}
}
    \caption{
    $R^2$ on both original and flatten data in SOTA ML-based method TimingGCN~\cite{guo2022timing}.
    $R^2_f$ on flatten data ($\mathbb{R}^{4n}$) is much higher than $R^2_{uf}$ on original data ($\mathbb{R}^{n \times 4}$).
    As a result, $R^2_f$ cannot reflect the real performance.
}
\label{tab:flatten-or-not}
\end{table}

\subsection{Experiments on Other Metrics}
For fairness and integrity, we also evaluate baselines and our PreRoutGNN with flatten $R_f^2$, which shown in Table~\ref{tab:main-results-random-seeds-flatten-R2-only} and Table~\ref{tab:auxiliary-tasks-flatten-R2-and-MSE}.
Our approach still outperforms other baselines.
However, we want to emphasize that compared to flatten $R^2_f$, un-flatten $R^2_{uf}$ is a more appropriate evaluation metric.
\begin{table*}[!t]
    \centering
    
    \adjustbox{width=0.8\linewidth}{
    \begin{tabular}{ll|ccccccccccccccc}
        \toprule

\multicolumn{2}{c|}{\multirow{2}{*}{Dataset}} & \multicolumn{5}{c}{flatten $R^2_f$ ($\uparrow$)}\\
\multicolumn{2}{c|}{} & GCNII & GAT & GINE & TimingGCN & PreRoutGNN\\
\midrule
\multirow{14}{*}{train} & BM64 & 0.941 $\pm$ 0.030 & 0.815 $\pm$ 0.080 & 0.898 $\pm$ 0.052 & 0.864 $\pm$ 0.131 & \textbf{0.997 $\pm$ 0.001}\\
~ & aes128 & 0.923 $\pm$ 0.035 & 0.852 $\pm$ 0.055 & 0.895 $\pm$ 0.031 & 0.889 $\pm$ 0.060 & \textbf{0.993 $\pm$ 0.001}\\
~ & aes256 & 0.936 $\pm$ 0.029 & 0.855 $\pm$ 0.057 & 0.905 $\pm$ 0.031 & 0.934 $\pm$ 0.021 & \textbf{0.997 $\pm$ 0.002}\\
~ & aes\_cipher & 0.908 $\pm$ 0.038 & 0.840 $\pm$ 0.060 & 0.849 $\pm$ 0.064 & 0.971 $\pm$ 0.017 & \textbf{0.998 $\pm$ 0.000}\\
~ & blabla & 0.946 $\pm$ 0.032 & 0.851 $\pm$ 0.066 & 0.877 $\pm$ 0.058 & 0.970 $\pm$ 0.010 & \textbf{0.998 $\pm$ 0.001}\\
~ & cic\_decimator & 0.861 $\pm$ 0.055 & 0.780 $\pm$ 0.102 & 0.770 $\pm$ 0.067 & 0.973 $\pm$ 0.012 & \textbf{0.998 $\pm$ 0.000}\\
~ & des & 0.978 $\pm$ 0.019 & 0.872 $\pm$ 0.071 & 0.941 $\pm$ 0.067 & 0.990 $\pm$ 0.004 & \textbf{0.999 $\pm$ 0.000}\\
~ & genericfir & 0.941 $\pm$ 0.028 & 0.700 $\pm$ 0.154 & 0.910 $\pm$ 0.042 & 0.846 $\pm$ 0.025 & \textbf{0.979 $\pm$ 0.001}\\
~ & picorv32a & 0.916 $\pm$ 0.037 & 0.822 $\pm$ 0.070 & 0.850 $\pm$ 0.052 & 0.950 $\pm$ 0.007 & \textbf{0.996 $\pm$ 0.001}\\
~ & salsa20 & 0.941 $\pm$ 0.026 & 0.828 $\pm$ 0.076 & 0.896 $\pm$ 0.037 & 0.941 $\pm$ 0.031 & \textbf{0.998 $\pm$ 0.001}\\
~ & usb & 0.865 $\pm$ 0.045 & 0.832 $\pm$ 0.066 & 0.780 $\pm$ 0.062 & 0.954 $\pm$ 0.026 & \textbf{0.998 $\pm$ 0.000}\\
~ & usb\_cdc\_core & 0.850 $\pm$ 0.063 & 0.690 $\pm$ 0.135 & 0.698 $\pm$ 0.106 & 0.967 $\pm$ 0.007 & \textbf{0.998 $\pm$ 0.001}\\
~ & wbqspiflash & 0.872 $\pm$ 0.052 & 0.789 $\pm$ 0.091 & 0.711 $\pm$ 0.107 & 0.967 $\pm$ 0.016 & \textbf{0.996 $\pm$ 0.002}\\
~ & zipdiv & 0.898 $\pm$ 0.035 & 0.863 $\pm$ 0.057 & 0.715 $\pm$ 0.151 & 0.971 $\pm$ 0.015 & \textbf{0.998 $\pm$ 0.001}\\
\midrule
\multirow{7}{*}{test} & aes192 & 0.876 $\pm$ 0.027 & 0.771 $\pm$ 0.036 & 0.867 $\pm$ 0.021 & 0.923 $\pm$ 0.026 & \textbf{0.996 $\pm$ 0.000}\\
~ & jpeg\_encoder & 0.829 $\pm$ 0.019 & 0.432 $\pm$ 0.132 & 0.778 $\pm$ 0.058 & 0.938 $\pm$ 0.019 & \textbf{0.988 $\pm$ 0.003}\\
~ & spm & 0.002 $\pm$ 0.266 & -0.168 $\pm$ 0.110 & 0.259 $\pm$ 0.152 & 0.848 $\pm$ 0.066 & \textbf{0.989 $\pm$ 0.005}\\
~ & synth\_ram & -0.426 $\pm$ 0.025 & -0.560 $\pm$ 0.018 & -0.499 $\pm$ 0.122 & 0.798 $\pm$ 0.057 & \textbf{0.984 $\pm$ 0.006}\\
~ & usbf\_device & 0.706 $\pm$ 0.019 & 0.549 $\pm$ 0.012 & 0.731 $\pm$ 0.039 & 0.934 $\pm$ 0.018 & \textbf{0.994 $\pm$ 0.000}\\
~ & xtea & 0.293 $\pm$ 0.044 & 0.199 $\pm$ 0.047 & 0.238 $\pm$ 0.100 & 0.886 $\pm$ 0.072 & \textbf{0.991 $\pm$ 0.001}\\
~ & y\_huff & 0.518 $\pm$ 0.231 & 0.451 $\pm$ 0.045 & 0.294 $\pm$ 0.380 & 0.914 $\pm$ 0.028 & \textbf{0.975 $\pm$ 0.011}\\
\midrule
\multicolumn{2}{c|}{Avg. train} & 0.913  & 0.813  & 0.835  & 0.942  & \textbf{0.996}\\
\multicolumn{2}{c|}{Avg. test} & 0.400  & 0.239  & 0.381  & 0.891  & \textbf{0.988}\\

\bottomrule
\end{tabular}
}
\caption{
    Slack prediction with flatten $R_{f}^2$ as metrics. 
    Experiments are implemented on 5 different seeds.
    }
\label{tab:main-results-random-seeds-flatten-R2-only}
\end{table*}
\begin{table*}[!t]
    \centering
    
    \adjustbox{width=1.0\linewidth}{
    \begin{tabular}{ll|cc|cc|cc|cc|cc|cc}
        \toprule

\multicolumn{2}{c|}{\multirow{2}{*}{Dataset}} & \multicolumn{2}{c|}{MSE slew ($\downarrow$)} & \multicolumn{2}{c|}{MSE net delay ($\downarrow$)} & \multicolumn{2}{c|}{MSE cell delay ($\downarrow$)} & \multicolumn{2}{c|}{flatten $R^2_f$ slew ($\uparrow$)} & \multicolumn{2}{c|}{flatten $R^2_f$ net delay ($\uparrow$)} & \multicolumn{2}{c}{flatten $R^2_f$ cell delay ($\uparrow$)}\\
\multicolumn{2}{c|}{} & TimingGCN & PreRoutGNN & TimingGCN & PreRoutGNN & TimingGCN & PreRoutGNN & TimingGCN & PreRoutGNN & TimingGCN & PreRoutGNN & TimingGCN & PreRoutGNN\\
\midrule
\multirow{14}{*}{train} & BM64 & 0.032 $\pm$ 0.013 & \textbf{0.006 $\pm$ 0.002} & 0.045 $\pm$ 0.022 & \textbf{0.022 $\pm$ 0.003} & 0.002 $\pm$ 0.001 & \textbf{0.000 $\pm$ 0.000} & 0.960 $\pm$ 0.016 & \textbf{0.993 $\pm$ 0.002} & 0.984 $\pm$ 0.008 & \textbf{0.992 $\pm$ 0.001} & 0.955 $\pm$ 0.030 & \textbf{0.995 $\pm$ 0.002}\\
~ & aes128 & 0.052 $\pm$ 0.018 & \textbf{0.005 $\pm$ 0.001} & 0.045 $\pm$ 0.009 & \textbf{0.021 $\pm$ 0.001} & 0.002 $\pm$ 0.001 & \textbf{0.000 $\pm$ 0.000} & 0.909 $\pm$ 0.032 & \textbf{0.991 $\pm$ 0.002} & 0.972 $\pm$ 0.006 & \textbf{0.987 $\pm$ 0.001} & 0.949 $\pm$ 0.022 & \textbf{0.996 $\pm$ 0.001}\\
~ & aes256 & 0.050 $\pm$ 0.015 & \textbf{0.006 $\pm$ 0.001} & 0.047 $\pm$ 0.011 & \textbf{0.023 $\pm$ 0.002} & 0.003 $\pm$ 0.002 & \textbf{0.000 $\pm$ 0.000} & 0.916 $\pm$ 0.025 & \textbf{0.990 $\pm$ 0.002} & 0.974 $\pm$ 0.006 & \textbf{0.988 $\pm$ 0.001} & 0.945 $\pm$ 0.029 & \textbf{0.996 $\pm$ 0.001}\\
~ & aes\_cipher & 0.048 $\pm$ 0.014 & \textbf{0.008 $\pm$ 0.000} & 0.061 $\pm$ 0.010 & \textbf{0.036 $\pm$ 0.003} & 0.002 $\pm$ 0.001 & \textbf{0.000 $\pm$ 0.000} & 0.922 $\pm$ 0.023 & \textbf{0.987 $\pm$ 0.000} & 0.971 $\pm$ 0.005 & \textbf{0.983 $\pm$ 0.002} & 0.970 $\pm$ 0.011 & \textbf{0.994 $\pm$ 0.000}\\
~ & blabla & 0.059 $\pm$ 0.024 & \textbf{0.011 $\pm$ 0.003} & 0.064 $\pm$ 0.029 & \textbf{0.029 $\pm$ 0.003} & 0.005 $\pm$ 0.003 & \textbf{0.001 $\pm$ 0.000} & 0.935 $\pm$ 0.027 & \textbf{0.988 $\pm$ 0.003} & 0.987 $\pm$ 0.006 & \textbf{0.994 $\pm$ 0.001} & 0.907 $\pm$ 0.057 & \textbf{0.987 $\pm$ 0.003}\\
~ & cic\_decimator & 0.028 $\pm$ 0.012 & \textbf{0.002 $\pm$ 0.001} & 0.016 $\pm$ 0.011 & \textbf{0.004 $\pm$ 0.000} & 0.001 $\pm$ 0.000 & \textbf{0.000 $\pm$ 0.000} & 0.955 $\pm$ 0.019 & \textbf{0.997 $\pm$ 0.002} & 0.987 $\pm$ 0.009 & \textbf{0.997 $\pm$ 0.000} & 0.976 $\pm$ 0.010 & \textbf{0.996 $\pm$ 0.001}\\
~ & des & 0.058 $\pm$ 0.024 & \textbf{0.005 $\pm$ 0.001} & 0.051 $\pm$ 0.011 & \textbf{0.023 $\pm$ 0.001} & 0.003 $\pm$ 0.002 & \textbf{0.000 $\pm$ 0.000} & 0.919 $\pm$ 0.034 & \textbf{0.993 $\pm$ 0.002} & 0.976 $\pm$ 0.005 & \textbf{0.989 $\pm$ 0.001} & 0.978 $\pm$ 0.015 & \textbf{0.998 $\pm$ 0.001}\\
~ & genericfir & 0.018 $\pm$ 0.006 & \textbf{0.004 $\pm$ 0.000} & 0.029 $\pm$ 0.007 & \textbf{0.018 $\pm$ 0.002} & 0.001 $\pm$ 0.001 & \textbf{0.000 $\pm$ 0.000} & 0.966 $\pm$ 0.012 & \textbf{0.991 $\pm$ 0.000} & 0.976 $\pm$ 0.006 & \textbf{0.986 $\pm$ 0.002} & 0.960 $\pm$ 0.022 & \textbf{0.998 $\pm$ 0.001}\\
~ & picorv32a & 0.093 $\pm$ 0.014 & \textbf{0.062 $\pm$ 0.002} & 0.051 $\pm$ 0.015 & \textbf{0.025 $\pm$ 0.003} & 0.002 $\pm$ 0.001 & \textbf{0.000 $\pm$ 0.000} & 0.879 $\pm$ 0.019 & \textbf{0.920 $\pm$ 0.003} & 0.982 $\pm$ 0.005 & \textbf{0.991 $\pm$ 0.001} & 0.964 $\pm$ 0.014 & \textbf{0.993 $\pm$ 0.002}\\
~ & salsa20 & 0.043 $\pm$ 0.012 & \textbf{0.007 $\pm$ 0.001} & 0.054 $\pm$ 0.013 & \textbf{0.030 $\pm$ 0.002} & 0.003 $\pm$ 0.002 & \textbf{0.000 $\pm$ 0.000} & 0.936 $\pm$ 0.018 & \textbf{0.989 $\pm$ 0.002} & 0.983 $\pm$ 0.004 & \textbf{0.991 $\pm$ 0.001} & 0.938 $\pm$ 0.028 & \textbf{0.992 $\pm$ 0.002}\\
~ & usb & 0.033 $\pm$ 0.016 & \textbf{0.002 $\pm$ 0.001} & 0.020 $\pm$ 0.014 & \textbf{0.004 $\pm$ 0.000} & 0.001 $\pm$ 0.001 & \textbf{0.000 $\pm$ 0.000} & 0.956 $\pm$ 0.022 & \textbf{0.997 $\pm$ 0.001} & 0.991 $\pm$ 0.007 & \textbf{0.998 $\pm$ 0.000} & 0.979 $\pm$ 0.013 & \textbf{0.997 $\pm$ 0.001}\\
~ & usb\_cdc\_core & 0.070 $\pm$ 0.011 & \textbf{0.045 $\pm$ 0.001} & 0.022 $\pm$ 0.010 & \textbf{0.008 $\pm$ 0.001} & 0.001 $\pm$ 0.001 & \textbf{0.000 $\pm$ 0.000} & 0.882 $\pm$ 0.018 & \textbf{0.925 $\pm$ 0.001} & 0.982 $\pm$ 0.009 & \textbf{0.993 $\pm$ 0.000} & 0.977 $\pm$ 0.011 & \textbf{0.996 $\pm$ 0.001}\\
~ & wbqspiflash & 0.043 $\pm$ 0.014 & \textbf{0.012 $\pm$ 0.002} & 0.030 $\pm$ 0.018 & \textbf{0.009 $\pm$ 0.001} & 0.002 $\pm$ 0.001 & \textbf{0.000 $\pm$ 0.000} & 0.937 $\pm$ 0.020 & \textbf{0.982 $\pm$ 0.002} & 0.990 $\pm$ 0.006 & \textbf{0.997 $\pm$ 0.000} & 0.970 $\pm$ 0.009 & \textbf{0.994 $\pm$ 0.001}\\
~ & zipdiv & 0.045 $\pm$ 0.012 & \textbf{0.020 $\pm$ 0.001} & 0.024 $\pm$ 0.019 & \textbf{0.004 $\pm$ 0.001} & 0.001 $\pm$ 0.000 & \textbf{0.000 $\pm$ 0.000} & 0.950 $\pm$ 0.013 & \textbf{0.978 $\pm$ 0.001} & 0.989 $\pm$ 0.009 & \textbf{0.998 $\pm$ 0.000} & 0.978 $\pm$ 0.008 & \textbf{0.997 $\pm$ 0.001}\\
\midrule
\multirow{7}{*}{test} & aes192 & 0.062 $\pm$ 0.012 & \textbf{0.011 $\pm$ 0.002} & 0.081 $\pm$ 0.018 & \textbf{0.037 $\pm$ 0.002} & 0.003 $\pm$ 0.002 & \textbf{0.000 $\pm$ 0.000} & 0.896 $\pm$ 0.021 & \textbf{0.981 $\pm$ 0.003} & 0.954 $\pm$ 0.010 & \textbf{0.979 $\pm$ 0.001} & 0.936 $\pm$ 0.031 & \textbf{0.992 $\pm$ 0.001}\\
~ & jpeg\_encoder & 0.071 $\pm$ 0.013 & \textbf{0.017 $\pm$ 0.000} & 0.097 $\pm$ 0.011 & \textbf{0.053 $\pm$ 0.002} & 0.004 $\pm$ 0.001 & \textbf{0.001 $\pm$ 0.000} & 0.882 $\pm$ 0.022 & \textbf{0.972 $\pm$ 0.000} & 0.966 $\pm$ 0.004 & \textbf{0.981 $\pm$ 0.001} & 0.924 $\pm$ 0.023 & \textbf{0.987 $\pm$ 0.001}\\
~ & spm & 0.036 $\pm$ 0.007 & \textbf{0.007 $\pm$ 0.000} & 0.066 $\pm$ 0.008 & \textbf{0.049 $\pm$ 0.016} & 0.002 $\pm$ 0.000 & \textbf{0.000 $\pm$ 0.000} & 0.945 $\pm$ 0.010 & \textbf{0.989 $\pm$ 0.001} & 0.921 $\pm$ 0.010 & \textbf{0.941 $\pm$ 0.019} & 0.946 $\pm$ 0.008 & \textbf{0.992 $\pm$ 0.000}\\
~ & synth\_ram & 0.155 $\pm$ 0.028 & \textbf{0.102 $\pm$ 0.047} & 0.268 $\pm$ 0.052 & \textbf{0.099 $\pm$ 0.004} & 0.012 $\pm$ 0.008 & \textbf{0.010 $\pm$ 0.005} & 0.827 $\pm$ 0.031 & \textbf{0.887 $\pm$ 0.052} & 0.929 $\pm$ 0.014 & \textbf{0.974 $\pm$ 0.001} & 0.881 $\pm$ 0.076 & \textbf{0.899 $\pm$ 0.047}\\
~ & usbf\_device & 0.059 $\pm$ 0.010 & \textbf{0.031 $\pm$ 0.002} & 0.055 $\pm$ 0.002 & \textbf{0.052 $\pm$ 0.006} & 0.004 $\pm$ 0.002 & \textbf{0.001 $\pm$ 0.000} & 0.910 $\pm$ 0.015 & \textbf{0.952 $\pm$ 0.003} & 0.968 $\pm$ 0.001 & \textbf{0.970 $\pm$ 0.003} & 0.927 $\pm$ 0.031 & \textbf{0.990 $\pm$ 0.002}\\
~ & xtea & 0.085 $\pm$ 0.035 & \textbf{0.011 $\pm$ 0.001} & 0.095 $\pm$ 0.019 & \textbf{0.064 $\pm$ 0.000} & 0.004 $\pm$ 0.002 & \textbf{0.001 $\pm$ 0.000} & 0.887 $\pm$ 0.046 & \textbf{0.986 $\pm$ 0.001} & 0.944 $\pm$ 0.011 & \textbf{0.963 $\pm$ 0.000} & 0.922 $\pm$ 0.043 & \textbf{0.989 $\pm$ 0.003}\\
~ & y\_huff & 0.076 $\pm$ 0.016 & \textbf{0.066 $\pm$ 0.000} & 0.211 $\pm$ 0.049 & \textbf{0.124 $\pm$ 0.016} & \textbf{0.003 $\pm$ 0.001} & 0.005 $\pm$ 0.003 & 0.915 $\pm$ 0.018 & \textbf{0.926 $\pm$ 0.000} & 0.940 $\pm$ 0.014 & \textbf{0.964 $\pm$ 0.005} & \textbf{0.953 $\pm$ 0.017} & 0.923 $\pm$ 0.042\\
\midrule
\multicolumn{2}{c|}{Avg. train} & 0.048 & \textbf{0.014} & 0.040 & \textbf{0.018} & 0.002 & \textbf{0.000} & 0.930 & \textbf{0.980} & 0.982 & \textbf{0.992} & 0.960 & \textbf{0.995}\\
\multicolumn{2}{c|}{Avg. test} & 0.078 & \textbf{0.035} & 0.125 & \textbf{0.068} & 0.005 & \textbf{0.003} & 0.894 & \textbf{0.956} & 0.946 & \textbf{0.967} & 0.927 & \textbf{0.968}\\

\bottomrule
\end{tabular}
}

\caption{
Comparison of other tasks, including prediction of slew, net delay and cell delay.
We evaluate baseline and our PreRoutGNN with metric flatten $R_f^2$ and MSE.
Experiments are implemented on five different random seeds.
    }

\label{tab:auxiliary-tasks-flatten-R2-and-MSE}
\end{table*}

We also evaluate different methods with Mean Squared Error (MSE) shown in Table~\ref{tab:main-results-large}.
\begin{table*}[!t]
    \centering
    
    \adjustbox{width=1.0\linewidth}
    { \small
    \begin{tabular}{ll|ccccc|cccccccccc}
        \toprule

\multicolumn{2}{c|}{\multirow{2}{*}{Dataset}} & \multicolumn{5}{c|}{MSE ($\downarrow$)} & \multicolumn{5}{c}{$R^2_{uf}$ ($\uparrow$)}\\
\multicolumn{2}{c|}{} & GCNII & GAT & GINE & TimingGCN & PreRoutGNN & GCNII & GAT & GINE & TimingGCN & PreRoutGNN\\
\midrule
\multirow{14}{*}{\rotatebox{90}{train}} & BM64 & 1.728 $\pm$ 0.881 & 5.442 $\pm$ 2.342 & 3.002 $\pm$ 1.544 & 3.997 $\pm$ 3.848 & \textbf{0.076 $\pm$ 0.038} & 0.763 $\pm$ 0.101 & 0.354 $\pm$ 0.223 & 0.623 $\pm$ 0.161 & 0.668 $\pm$ 0.252 & \textbf{0.986 $\pm$ 0.006}\\
~ & aes128 & 3.195 $\pm$ 1.426 & 6.096 $\pm$ 2.282 & 4.346 $\pm$ 1.279 & 4.581 $\pm$ 2.470 & \textbf{0.307 $\pm$ 0.024} & 0.458 $\pm$ 0.207 & -0.245 $\pm$ 0.340 & 0.029 $\pm$ 0.352 & -0.055 $\pm$ 0.312 & \textbf{0.822 $\pm$ 0.008}\\
~ & aes256 & 3.119 $\pm$ 1.408 & 7.078 $\pm$ 2.778 & 4.617 $\pm$ 1.489 & 3.238 $\pm$ 1.037 & \textbf{0.169 $\pm$ 0.079} & 0.581 $\pm$ 0.151 & -0.097 $\pm$ 0.306 & 0.067 $\pm$ 0.311 & 0.499 $\pm$ 0.102 & \textbf{0.937 $\pm$ 0.023}\\
~ & aes\_cipher & 1.391 $\pm$ 0.569 & 2.408 $\pm$ 0.899 & 2.269 $\pm$ 0.958 & 0.429 $\pm$ 0.255 & \textbf{0.034 $\pm$ 0.007} & 0.020 $\pm$ 0.447 & -0.438 $\pm$ 0.427 & -0.162 $\pm$ 0.242 & 0.644 $\pm$ 0.123 & \textbf{0.967 $\pm$ 0.000}\\
~ & blabla & 6.123 $\pm$ 3.651 & 17.083 $\pm$ 7.588 & 14.044 $\pm$ 6.629 & 3.475 $\pm$ 1.125 & \textbf{0.233 $\pm$ 0.076} & 0.889 $\pm$ 0.073 & 0.679 $\pm$ 0.131 & 0.737 $\pm$ 0.122 & 0.958 $\pm$ 0.013 & \textbf{0.997 $\pm$ 0.001}\\
~ & cic\_decimator & 0.468 $\pm$ 0.184 & 0.738 $\pm$ 0.342 & 0.770 $\pm$ 0.225 & 0.091 $\pm$ 0.040 & \textbf{0.006 $\pm$ 0.000} & 0.719 $\pm$ 0.112 & 0.597 $\pm$ 0.163 & 0.589 $\pm$ 0.107 & 0.963 $\pm$ 0.014 & \textbf{0.996 $\pm$ 0.000}\\
~ & des & 1.866 $\pm$ 1.662 & 11.066 $\pm$ 6.096 & 5.129 $\pm$ 5.805 & 0.856 $\pm$ 0.355 & \textbf{0.055 $\pm$ 0.004} & 0.924 $\pm$ 0.056 & 0.566 $\pm$ 0.225 & 0.776 $\pm$ 0.220 & 0.905 $\pm$ 0.058 & \textbf{0.993 $\pm$ 0.001}\\
~ & genericfir & 0.234 $\pm$ 0.112 & 1.190 $\pm$ 0.613 & 0.359 $\pm$ 0.168 & 0.612 $\pm$ 0.099 & \textbf{0.085 $\pm$ 0.005} & 0.710 $\pm$ 0.122 & -0.436 $\pm$ 0.733 & 0.503 $\pm$ 0.219 & 0.222 $\pm$ 0.123 & \textbf{0.871 $\pm$ 0.013}\\
~ & picorv32a & 2.346 $\pm$ 1.033 & 4.952 $\pm$ 1.957 & 4.167 $\pm$ 1.440 & 1.386 $\pm$ 0.189 & \textbf{0.109 $\pm$ 0.028} & 0.672 $\pm$ 0.096 & 0.239 $\pm$ 0.240 & 0.425 $\pm$ 0.169 & 0.839 $\pm$ 0.045 & \textbf{0.974 $\pm$ 0.007}\\
~ & salsa20 & 3.991 $\pm$ 1.751 & 11.635 $\pm$ 5.117 & 7.024 $\pm$ 2.475 & 4.005 $\pm$ 2.107 & \textbf{0.160 $\pm$ 0.075} & 0.814 $\pm$ 0.066 & 0.530 $\pm$ 0.170 & 0.717 $\pm$ 0.073 & 0.792 $\pm$ 0.118 & \textbf{0.994 $\pm$ 0.002}\\
~ & usb & 0.580 $\pm$ 0.192 & 0.725 $\pm$ 0.285 & 0.947 $\pm$ 0.267 & 0.197 $\pm$ 0.112 & \textbf{0.008 $\pm$ 0.001} & 0.747 $\pm$ 0.088 & 0.678 $\pm$ 0.115 & 0.607 $\pm$ 0.118 & 0.934 $\pm$ 0.035 & \textbf{0.996 $\pm$ 0.000}\\
~ & usb\_cdc\_core & 0.830 $\pm$ 0.348 & 1.713 $\pm$ 0.745 & 1.673 $\pm$ 0.588 & 0.185 $\pm$ 0.041 & \textbf{0.013 $\pm$ 0.003} & 0.793 $\pm$ 0.095 & 0.502 $\pm$ 0.212 & 0.563 $\pm$ 0.177 & 0.961 $\pm$ 0.008 & \textbf{0.997 $\pm$ 0.001}\\
~ & wbqspiflash & 1.321 $\pm$ 0.532 & 2.181 $\pm$ 0.936 & 2.988 $\pm$ 1.106 & 0.343 $\pm$ 0.169 & \textbf{0.039 $\pm$ 0.021} & 0.742 $\pm$ 0.091 & 0.613 $\pm$ 0.131 & 0.522 $\pm$ 0.136 & 0.956 $\pm$ 0.013 & \textbf{0.994 $\pm$ 0.002}\\
~ & zipdiv & 1.307 $\pm$ 0.442 & 1.750 $\pm$ 0.725 & 3.645 $\pm$ 1.939 & 0.370 $\pm$ 0.187 & \textbf{0.027 $\pm$ 0.014} & 0.822 $\pm$ 0.070 & 0.703 $\pm$ 0.129 & 0.593 $\pm$ 0.188 & 0.971 $\pm$ 0.011 & \textbf{0.998 $\pm$ 0.001}\\
\midrule
\multirow{7}{*}{\rotatebox{90}{test}} & aes192 & 6.429 $\pm$ 1.407 & 11.918 $\pm$ 1.889 & 6.934 $\pm$ 1.105 & 4.000 $\pm$ 1.328 & \textbf{0.228 $\pm$ 0.015} & -0.017 $\pm$ 0.188 & -0.727 $\pm$ 0.173 & -0.048 $\pm$ 0.352 & 0.384 $\pm$ 0.205 & \textbf{0.937 $\pm$ 0.009}\\
~ & jpeg\_encoder & 10.893 $\pm$ 1.237 & 36.278 $\pm$ 8.409 & 14.163 $\pm$ 3.730 & 3.950 $\pm$ 1.208 & \textbf{0.747 $\pm$ 0.209} & 0.056 $\pm$ 0.080 & -2.859 $\pm$ 0.667 & -0.285 $\pm$ 0.198 & 0.478 $\pm$ 0.131 & \textbf{0.764 $\pm$ 0.010}\\
~ & spm & 1.444 $\pm$ 0.384 & 1.690 $\pm$ 0.159 & 1.071 $\pm$ 0.220 & 0.220 $\pm$ 0.096 & \textbf{0.016 $\pm$ 0.007} & -0.247 $\pm$ 0.309 & -0.474 $\pm$ 0.134 & 0.040 $\pm$ 0.161 & 0.802 $\pm$ 0.083 & \textbf{0.985 $\pm$ 0.006}\\
~ & synth\_ram & 65.672 $\pm$ 1.137 & 71.823 $\pm$ 0.828 & 69.009 $\pm$ 5.632 & 9.302 $\pm$ 2.617 & \textbf{0.753 $\pm$ 0.260} & -0.913 $\pm$ 0.031 & -1.078 $\pm$ 0.025 & -1.007 $\pm$ 0.162 & 0.735 $\pm$ 0.079 & \textbf{0.977 $\pm$ 0.008}\\
~ & usbf\_device & 5.545 $\pm$ 0.365 & 8.497 $\pm$ 0.226 & 5.062 $\pm$ 0.743 & 1.249 $\pm$ 0.339 & \textbf{0.117 $\pm$ 0.001} & -0.215 $\pm$ 0.065 & -0.946 $\pm$ 0.146 & -0.133 $\pm$ 0.094 & 0.347 $\pm$ 0.119 & \textbf{0.949 $\pm$ 0.002}\\
~ & xtea & 22.387 $\pm$ 1.400 & 25.368 $\pm$ 1.489 & 24.132 $\pm$ 3.173 & 3.613 $\pm$ 2.270 & \textbf{0.285 $\pm$ 0.041} & 0.210 $\pm$ 0.044 & -0.055 $\pm$ 0.091 & 0.099 $\pm$ 0.181 & 0.888 $\pm$ 0.062 & \textbf{0.985 $\pm$ 0.000}\\
~ & y\_huff & 4.259 $\pm$ 2.036 & 4.849 $\pm$ 0.394 & 6.235 $\pm$ 3.354 & 0.764 $\pm$ 0.245 & \textbf{0.225 $\pm$ 0.099} & -0.235 $\pm$ 0.450 & -0.481 $\pm$ 0.114 & -1.034 $\pm$ 0.999 & 0.490 $\pm$ 0.146 & \textbf{0.931 $\pm$ 0.029}\\
\midrule
\multicolumn{2}{c|}{Avg. train} & 2.036  & 5.290  & 3.927  & 1.698  & \textbf{0.095} & 0.689  & 0.303  & 0.471  & 0.733  & \textbf{0.966}\\
\multicolumn{2}{c|}{Avg. test} & 16.661  & 22.917  & 18.087  & 3.300  & \textbf{0.339} & -0.195  & -0.946  & -0.338  & 0.589  & \textbf{0.933}\\

\bottomrule
\end{tabular}
}

\caption{
    Slack prediction with MSE and $R_{uf}^2$ as primary metrics. 
    Experiments are implemented on 5 different seeds.
}

\label{tab:main-results-large}

\end{table*}

\subsection{Slack Prediction on Endpoints}
\begin{figure*}[!tb]
    \centering
    \includegraphics[width=1.0\linewidth]{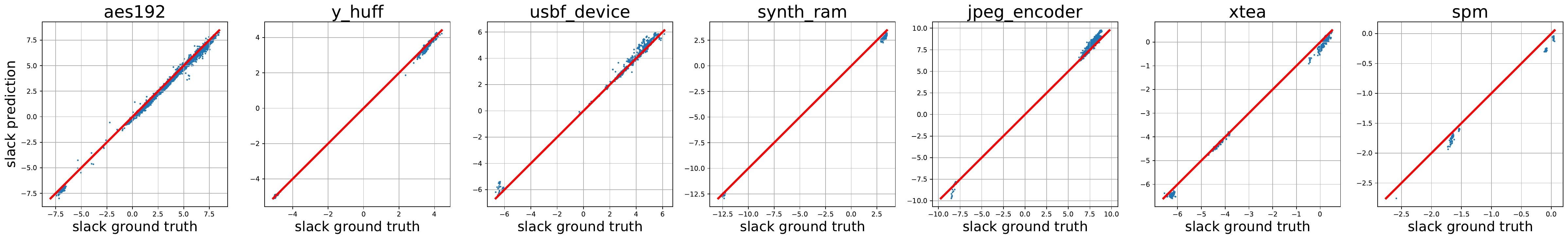}
    \caption{Slack ground truth and prediction on timing path endpoints, showing a strong linear correlation (red line is $y = x$).}
    \label{fig:slack-prediction}
\end{figure*}

We draw slack ground truth and prediction on all timing endpoints of testing set in Fig.~\ref{fig:slack-prediction}, which shows a strong linear correlation between prediction and ground truth.

\subsection{Time Cost for Graph Partition and Inference}
\begin{figure}[!tb]
    \centering
    \includegraphics[width=1.0\linewidth]{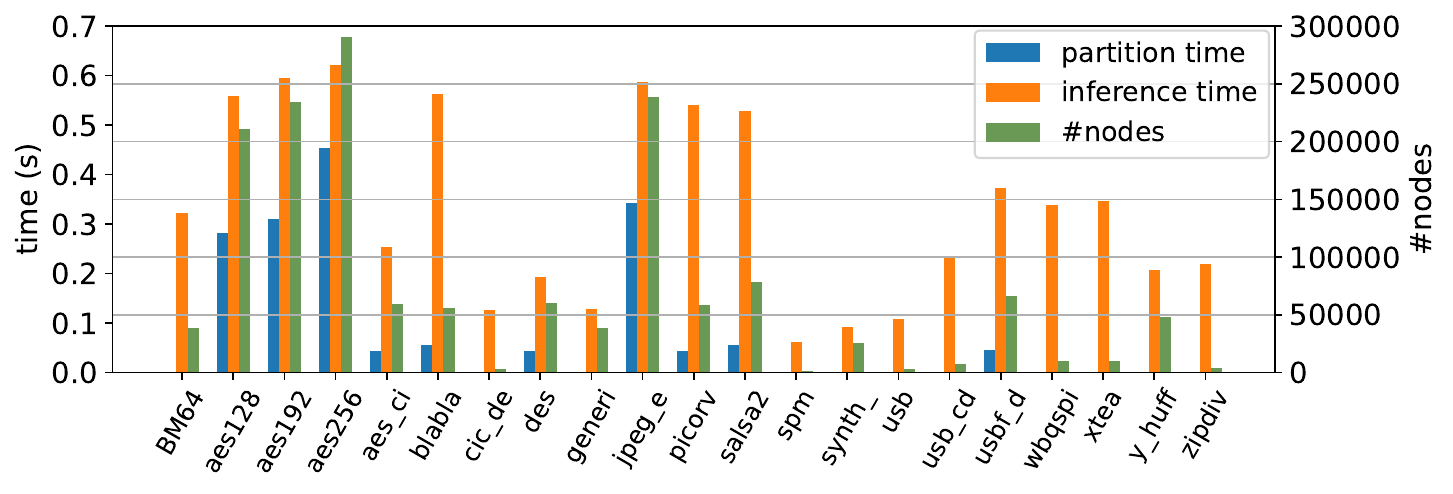}
    \caption{
    Partition time with maximum sub-graph size 50000.
    If the size of original graph is less than maximum sub-graph size, we will not partition this circuit.
    }
    \label{fig:partition_time-inference_time}
\end{figure}

We also evaluate the time cost for graph partition and inference shown in Fig.~\ref{fig:partition_time-inference_time}.
For large-scale circuit with about 300,000 pins, the inference can be finished within one second.

\section{Why Other Graph Partition Algorithms Inappropriate?}
As we have discussed, the AT of current node relies on \emph{all} its predecessors.
We can and only can calculate its AT after \emph{all} ATs of its predecessors have been calculated.
We apply topological sorting to depict this dependency.

In our graph partition algorithm, we add nodes belonging to the same topological sorting order into a level, and construct sub-graph based on continuous levels. 
If we apply topological sorting in the sub-graph, the relative sorting orders will be the same in the original whole graph, which preserve the topological dependency in timing path.

\begin{figure}[!t]
    \centering
    \includegraphics[width=1.0\linewidth]{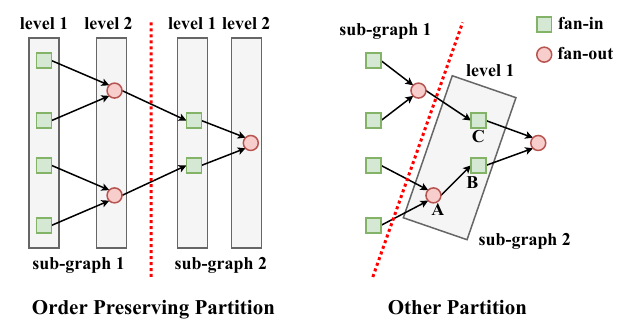}
    \caption{
    Our order preserving partition algorithm, ensuring that after partitioning, nodes belonging to the same topological sorting order serve the same logic role (all of them are fan-in or fan-out), and they are still in the same level.
    As a result, our graph partition algorithm can preserve the sorting order.
    }
    \label{fig:other-partition}
\end{figure}

Traditional graph clustering methods~\cite{hartigan1979algorithm} merge nodes into a cluster, leading to the loss of fine-grained connections. 
In this task, we need to calculate AT of each node.
As a result, it is inappropriate in this settings.
Other partitioning methods, such as KaHIP~\cite{sanders2013think} and METIS~\cite{karypis1998software} are based on the optimizing for cut- or connectivity-metric, mainly focus on neighbours structure, while in circuit, deep timing paths play a critical role. 
Furthermore, in topological propagation layer, the relative topological sorting orders will be different from that in the original whole graph, which disrupts the topological sorting level information in circuit graph, even makes propagation unable to implement.
For example, in original whole graph and our partitioned sub-graphs, nodes belonging to the same order serve as the same logic role, which means they are all fan-in or fan-out.
However, if we apply topological sorting on sub-graphs partitioned by other algorithms, nodes belonging to the same order possibly do not serve the same logic role.

We also illustrate it in Fig~\ref{fig:other-partition}.
Green rectangle means fan-in, which is input pin of cell.
Red circle means fan-out, which is output pin of cell.
Under usual circumstances, a cell consists with two input pins and a single output pin.
The dotted line splits the whole graph into two sub-graphs.
Left is our order preserving partition algorithm, ensuring that after partitioning, nodes belonging to the same topological sorting order still serve as the same logic role (all of them are fan-in or fan-out), and they are still in the same level.
However, in the right part, results through other partition algorithms, possibly are faced with such situation.
After applying topological sorting on the sub-graph, nodes belonging to the same topological order do not always serve as the same logic role.
For example, in sub-graph 2 of other graph partition, A has the same topological sorting order with B and C, but A serves as a fan-out, which is the output pin of a cell, while B and C serve as fan-in, which are the input pin of another cell.
This makes topological propagation unable to implement.
As a result, our graph partition algorithm can preserve the sorting order.

\section{Model Structure}
\subsection{Global Pre-training Circuit Graph Auto-Encoder}
The forward propagation of our GNN layer is shown as follows:
\begin{equation}
    \begin{aligned}
        \mathbf{m}_{ji} &= \mathrm{MLP}(\mathbf{f}_j || \mathbf{f}_i || \mathbf{e}_{ji}) + \mathbf{f}_j \\
        \mathbf{a}_i &=  \frac{\sum_{j \in \mathcal{N}(i)} \mathbf{m}_{ji}}{\mathrm{size}(\mathcal{N}(i))} \\
        \mathbf{b}_i &= \max_{j \in \mathcal{N}(i)} \mathbf{m}_{ji} \\
        \mathbf{c}_i &= \min_{j \in \mathcal{N}(i)} \mathbf{m}_{ji} \\
        \mathbf{d}_i &= \sum_{j \in \mathcal{N}(i)} \mathbf{m}_{ji} \\
        \mathbf{f}_i &= \mathrm{MLP}(\mathbf{f}_i|| \mathbf{a}_i || \mathbf{b}_i || \mathbf{c}_i || \mathbf{d}_i) + \mathbf{f}_i,
    \end{aligned}
\end{equation}
where $\mathbf{f}_i$ is the node feature for node $i$, $\mathcal{N}(i)$ is the predecessors of $i$, $\mathbf{e}_{ji}$ is the edge feature for edge $(j,i)$.
We select 4 as number of encoder layer, 4 as number of decoder layer, 32 as hidden dimension and 4 as dimension of latent space. 
Both MLPs are 2 layers with 32 as hidden dimension. 
In MLP, we use leaky ReLU with negative slope 0.2 as activation.

\subsection{RLL-based GCN Layer}
We stack 4 layers of RLL-based GCN layers. 
Each MLP in it is 2 layers with 64 as hidden dimension. 
In MLP, we use leaky ReLU with negative slope 0.2 as activation.

We update node features in RLL-based GCN layer for net connection (edge type is `net' or `net\_inv') as follows:
\begin{equation}
    {
    \small
    \begin{aligned}
        \mathbf{m}_{ji} &= \mathrm{MLP}(\mathbf{f}_j || \mathbf{f}_i || \mathbf{e}_{ji}) + \mathbf{f}_j \\
        \mathbf{a}_i &=  \frac{\sum_{j \in \mathcal{N}(i)} \mathbf{m}_{ji}}{\mathrm{size}(\mathcal{N}(i))}, \mathbf{b}_i = \max_{j \in \mathcal{N}(i)} \mathbf{m}_{ji} \\
        \mathbf{f}_i &= \mathrm{MLP}(\mathbf{f}_i|| \mathbf{a}_i || \mathbf{b}_i) + \mathbf{f}_i,
    \end{aligned}
    }
\end{equation}
where $\mathbf{f}_i$ is the node feature for node $i$, $\mathcal{N}(i)$ is the predecessors of $i$, $\mathbf{e}_{ji}$ is the edge feature for edge $(j,i)$.

\subsection{Baselines}
For GCNII~\cite{chen2020simple}, we stack 8 layers with hidden dimension 64. We use leaky ReLU with negative slope 0.2 as activation.

For GAT~\cite{vaswani2017attention}, we stack 8 layers with hidden dimension 64. We use leaky ReLU with negative slope 0.2 as activation.

For GINE~\cite{hu2019strategies}, we stack 8 layers with node feature hidden dimension 64, edge feature hidden dimension 64. We use leaky ReLU with negative slope 0.2 as activation.

For TimingGCN, we use the same model structure in line with ~\cite{guo2022timing}.

\end{document}